\documentclass[final,3p,times]{elsarticle}

\usepackage{amssymb}
\usepackage{amsthm}

\usepackage{lineno}

\usepackage{multirow}
\usepackage{float}
\usepackage[ruled,vlined,linesnumbered]{algorithm2e}

\usepackage{amsmath}
\usepackage{threeparttable}

\journal{Image and Vision Computing}

\begin{document}

\begin{frontmatter}



    \title{Comparison of fine-tuning strategies for transfer learning in medical image classification}


\author[inst1]{Ana Davila\corref{cor1}}

\cortext[cor1]{Corresponding author}
\ead{davila.ana@robo.mein.nagoya-u.ac.jp}

\affiliation[inst1]{organization={Institutes of Innovation for Future Society, Nagoya University},
            addressline={Furo-cho, Chikusa-ku}, 
            city={Nagoya},
            postcode={464-8601}, 
            state={Aichi},
            country={Japan}}

\author[inst2]{Jacinto Colan}
\author[inst1]{Yasuhisa Hasegawa}

\affiliation[inst2]{organization={Department of Micro-Nano Mechanical Science and Engineering, Nagoya University},
            addressline={Furo-cho, Chikusa-ku}, 
            city={Nagoya},
            postcode={464-8603}, 
            state={Aichi},
            country={Japan}}

\begin{abstract}
In the context of medical imaging and machine learning, one of the most pressing challenges is the effective adaptation of pre-trained models to specialized medical contexts. Despite the availability of advanced pre-trained models, their direct application to the highly specialized and diverse field of medical imaging often falls short due to the unique characteristics of medical data. This study provides a comprehensive analysis on the performance of various fine-tuning methods applied to pre-trained models across a spectrum of medical imaging domains, including X-ray, MRI, Histology, Dermoscopy, and Endoscopic surgery. We evaluated eight fine-tuning strategies, including standard techniques such as fine-tuning all layers or fine-tuning only the classifier layers, alongside methods such as gradually unfreezing layers, regularization based fine-tuning and adaptive learning rates. We selected three well-established CNN architectures (ResNet-50, DenseNet-121, and VGG-19) to cover a range of learning and feature extraction scenarios. Although our results indicate that the efficacy of these fine-tuning methods significantly varies depending on both the architecture and the medical imaging type, strategies such as combining Linear Probing with Full Fine-tuning resulted in notable improvements in over 50\% of the evaluated cases, demonstrating general effectiveness across medical domains. Moreover, Auto-RGN, which dynamically adjusts learning rates, led to performance enhancements of up to 11\% for specific modalities. Additionally, the DenseNet architecture showed more pronounced benefits from alternative fine-tuning approaches compared to traditional full fine-tuning. This work not only provides valuable insights for optimizing pre-trained models in medical image analysis but also suggests the potential for future research into more advanced architectures and fine-tuning methods.
\end{abstract}


\end{frontmatter}


\section{Introduction}

Medical image analysis involves the extraction of vital information from medical images for diagnostic and therapeutic purposes. While advances in imaging technologies facilitate early detection and precise disease identification, manual analysis remains labor-intensive and error-prone, leading to inconsistent interpretations \cite{hu23reinforcement}. To overcome these challenges, automated techniques, particularly those employing advanced machine learning algorithms, have gained popularity for improving efficiency and reliability in medical image analysis \cite{shen17deep, litjens17survey}. A significant challenge in this field is the scarcity of large, annotated datasets, which are generally necessary for training robust, high-performing models. Transfer learning (TL) addresses this challenge by utilizing models trained on extensive, diverse datasets and applying them to specific medical contexts \cite{romero20targeted}.

Transfer learning is a versatile approach in deep learning. It applies knowledge from a source domain, rich in annotated instances, to a related but distinct target domain lacking sufficient labeled data \cite{pan10survey}. It is efficient in terms of resource utilization, saving time and effort that would otherwise be spent on data collection and labeling in the target domain. This method accelerates the training process by using pre-trained models as a starting point \cite{hussain19study}. In medical image analysis, TL has demonstrated its value by providing quality decision support and reducing reliance on extensive labeled medical datasets, which are often costly and time-consuming to gather \cite{kora22transfer, kim22transfer, peng22rethinking}. By using pre-trained models, TL leverages previously learned generic features from various image types, enabling more effective feature extraction from medical images. Popular TL models in this domain include AlexNet, ResNet, VGGNet, and DenseNet, which have proven effective in tasks such as segmentation \cite{sanford20data}, object identification \cite{koskinen22automated, lavanchy21automation}, surgical workflow analysis \cite{yamada23task, zhang20automatic}, and disease categorization \cite{manokaran21detection}.

However, the success of TL heavily depends on how closely the source and target domains align \cite{wang18deep}. Discrepancies, especially distribution shifts, can significantly affect the performance of TL models, sometimes leading to negative transfer \cite{recht19doimagenet}. These shifts occur when the data distributions in the source and target domains are dissimilar \cite{quinonero08dataset}. Medical images are particularly susceptible to these shifts due to natural variations arising from different factors such as scene composition, object types, and lighting conditions \cite{taori20measuring}. In surgical settings, for instance, a variety of shapes, colors, and sizes of tools are present in endoscopic images \cite{fozilov23endoscope, colan23openrst}. A model trained on one type of surgical tool may not perform well with images of that tool in different surgical procedures. Similar challenges are present in other fields, such as image classification \cite{bhojanapalli21understanding}, continual localization \cite{cai21online}, protein characterization \cite{xu22improved, davila22abadapt}, and image-text association \cite{radford21learning}. Such distribution shifts can significantly affect the reliability of TL models.

Previous research has addressed the challenges of distribution shifts and negative transfer by developing robust and adaptable strategies that minimize reliance on labeled data in the target domain \cite{peters16causal, arjovsky19invariant}. Inductive transfer learning strategies, in contrast, use some labeled data from the target domain to improve the model’s accuracy and adaptability for the specific task. This approach is cost-effective and often outperforms domain generalization and unsupervised adaptation methods \cite{rosenfeld22domain, kirichenko22last}. Two main methods are highlighted: multi-task learning and sequential learning. Multi-task learning aims to learn multiple tasks simultaneously by discovering shared latent features, requiring balanced data distribution from both source and target domains during training. Sequential learning involves pre-training to develop a general model using extensive data from diverse tasks, followed by adaptation to the target task by fine-tuning pre-trained parameters for specificity.

\begin{figure}[H]
	\centering
	\includegraphics[width=0.5 \textwidth]{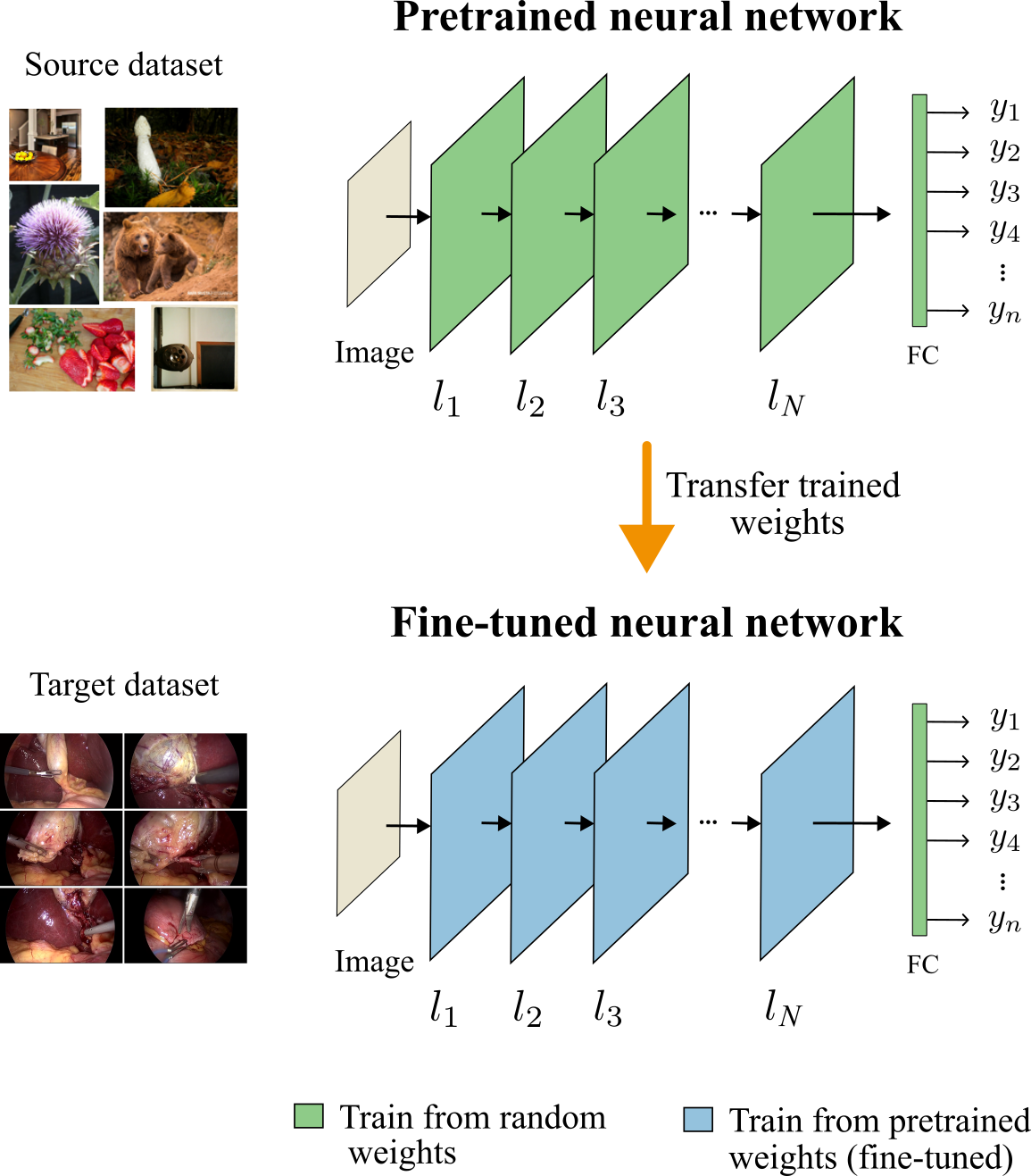}
	\caption{Schematic representation of the transfer learning process, illustrating the knowledge transfer from a pre-trained source model to a target medical imaging dataset.}
	\label{fig:1}
\end{figure}

Adaptation in transfer learning involves two primary techniques: feature extraction and fine-tuning \cite{zhuang21comprehensive}. Feature extraction uses fixed pre-trained weights to extract relevant features for the target task, coupled with a task-specific classifier \cite{shi19deep, nogueira17towards}. This method is known for its computational efficiency and potential for model reuse. Fine-tuning, on the other hand, adjusts pre-trained models to fit the target task, refining parameters based on specific task requirements, as shown in Fig.~\ref{fig:1}. It is a straightforward approach that allows a single model to adapt to various tasks with minimal adjustments.

Several fine-tuning strategies have been proposed. Full fine-tuning, which adjusts all layers of a pre-trained model for the new task, has been a common approach. However, recent studies suggest that this method may not always be optimal, especially when significant differences exist between source and target domain distributions, potentially leading to negative transfer \cite{kumar22fine}. To counter these challenges, various fine-tuning strategies have been proposed. One category involves exploration methods that navigate the parameter space guided by a cost function. Within this category, metaheuristic techniques allow selective fine-tuning of layers, exploring parameter combinations to enhance performance \cite{vrbancic20transfer, nagae22automatic}. These techniques can be effective but are computationally demanding and require precise definition of the search space. Alternatively, non-iterative fine-tuning strategies offer a more efficient approach by adjusting parameters without exhaustive exploration. These strategies include linear probing, training only the last classifier layer \cite{kumar22fine}, regularization that controls deviation from pre-trained parameters \cite{li18explicit}, selective freezing of layers for targeted tuning \cite{howard18universal, guo19spottune, mukherjee20distilling, davila23gradient}, and adapting learning rates for specific layers or groups \cite{shen21partial, ro21autolr, lee23surgical}.

Recognizing the potential of fine-tuning in medical image analysis, our study aims to bridge the gap in research by conducting a comprehensive comparison of non-iterative fine-tuning strategies in medical image classification. These methods are especially appealing in medical imaging due to their efficiency in avoiding multiple training iterations, which are resource-intensive. We have evaluated a range of fine-tuning strategies across various publicly available medical datasets from different medical domains. Each domain presents unique challenges commonly encountered in medical image classification, such as class imbalance, limited dataset size, or multi-label scenarios. To our knowledge, such a broad evaluation has not been previously undertaken. Our extensive comparison aims to assess the overall effectiveness and robustness of these strategies in medical image classification. The contributions of this work include:

\begin{itemize}
\item A review of fine-tuning methods applied to medical image classification tasks.
\item An analysis and comparison of classification performance across diverse medical imaging domains, such as X-ray, MRI, histology, dermoscopy, and endoscopy.
\item The identification of optimal fine-tuning strategies for specific architectures and domains.
\end{itemize}

The paper is organized as follows: Section~\ref{sec:2} provides a review of related work in fine-tuning methods for medical image classification. Section~\ref{sec:3} describes the fine-tuning strategies under review and the experimental setup, including dataset specifics, model architectures, evaluation metrics, and implementation details. Section~\ref{sec:4} presents a comparative analysis of the performance of these strategies. Section~\ref{sec:5} discusses the implications of our findings. The paper concludes with Section~\ref{sec:6}, summarizing the key insights and suggesting directions for future research.

\section{Fine-tuning for medical image classification} 
\label{sec:2} 
Fine-tuning can improve the performance of models and reduce the need for extensive training and data annotation, and has been widely used in the classification of medical images, where data is often scarce, variable, and complex. However, fine-tuning also poses some challenges in medical image classification. One of them is choosing a suitable network architecture that balances complexity and accuracy. The network architecture, especially its convolutional and pooling layers, also known as the backbone, affects the number of parameters, the computational cost, and the data representation capability of the network. Larger architectures, such as ResNet, GoogleNet, and DenseNet, can extract complex features but require more data and computing resources for fine-tuning. Smaller architectures, such as MobileNet and ShuffleNet, are more efficient and less prone to overfitting, but they may have lower accuracy on some tasks. The optimal architecture depends on the specific application, the available data, and the computational constraints and can vary between different studies. Another challenge is selecting a relevant source dataset for transfer learning. The source dataset should have high quality and similarity to the target dataset so that the learned features are transferable. ImageNet, a large dataset with diverse object classes, is commonly used as a source dataset, but it may not always match the specific domain requirements. Moreover, the different medical imaging modalities, such as X-rays, PET scans, ultrasound, MRIs, and microscopies, pose an additional challenge. Each modality has unique characteristics that influence image processing and classification. Developing models that can handle such diversity efficiently remains a difficult task.

This study focuses on fine-tuning techniques within five distinct medical imaging domains: X-rays, MRI, histology, dermoscopy, and endoscopy, which serve as representative modalities. Key studies in these areas are systematically summarized in Table~\ref{tab:1}. Detailed discussions of these medical domains and the corresponding applied fine-tuning strategies are presented in the following sections. These strategies are categorized as follows:

\begin{itemize}
    \item Full fine-tuning (FT): All layers of the pre-trained model are retrained.
    \item Linear Probing (LP): Only the classifier layers are retrained.
    \item Selective fine-tuning: Specific parameters, such as fine-tuning layers or learning rate, are selected and predetermined for retraining.
    \item Dynamic fine-tuning: Parameters are adjusted adaptively during the fine-tuning process.
\end{itemize}

\begin{table}[H]
\centering
    \caption{Overview of fine-tuning approaches for medical image classification }\label{tab:1}
    \scalebox{0.85}{
        \begin{tabular}{l c c c c c c}
            \hline
            \bf{Reference}  &\bf{Medical domain}  &\bf{Backbone}   &\bf{Source domain}	 &\bf{Fine-tune method}  &\bf{Application}\\
            \hline            
            Kermany et al. \cite{kermany18identifying}.         & X-ray         & Inception             & ImageNet      & LP            & Pneumonia \\
            Yadav et al. \cite{yadav19deep}                     & X-ray         & VGG, Inception        & ImageNet      & LP            & Pneumonia \\
            Liang et al. \cite{liang20transfer}                 & X-ray         & CNN                   & ChestX-ray14  & FT            & Penumonia     \\
            Chouhan et al. \cite{chouhan20novel}                & X-ray         & Ensemble              & ImageNet      & LP            & Pneumonia     \\
            Apostolopoulos et al. \cite{apostolopoulos20covid}  & X-ray         & Multiple              & ImageNet      & Selective     & Covid-19 \\
            Maghdid et al. \cite{maghdid20diagnosing}           & X-ray/CT      & AlexNet               & ImageNet      & FT            & Covid-19    \\
            Ahuja et al. \cite{ahuja21deep}                     & CT            & ResNet, SqueezeNet    & ImageNet      & FT            & Covid-19     \\
            Basaia et al. \cite{basaia19automated}              & MRI           & CNN                   & ADNI          & FT            & Alzheimer     \\
            Oh et al. \cite{oh19classification}                 & MRI           & Inception             & CAE           & FT            & Alzheimer     \\
            Eitel et al. \cite{eitel19uncovering}               & MRI           & CNN                   & ADNI          & FT            & Multiple sclerosis    \\
            Jonsson et al. \cite{jonsson19brain}                & MRI           & ResNet                & Own           & LP-FT         & Brain age \\
            Naser et al. \cite{naser20brain}                    & MRI           & VGG                   & TCIA          & LP            & Tumor grading     \\
            Nawaz et al. \cite{nawaz18classification}           & Microscopy    & AlexNet               & ImageNet      & LP            & Breast cancer \\
            Ferreira et al. \cite{ferreira18classification}     & Microscopy    & Inception Resnet      & ImageNet      & LP + Selective  & Breast cancer\\
            Bayramoglu \cite{bayramoglu16transfer}              & Microscopy    & Multiple              & ImageNet      & Dynamic       & Cell Nuclei  \\
            Vesal et al. \cite{vesal18classification}           & Microscopy    & Inception, ResNet     & ImageNet      & FT            & Breast cancer\\
            Vrbancic et al. \cite{vrbancic20transfer}           & Microscopy    & VGG                   & ImageNet      & Selective     & Osteosarcoma      \\
            Mahbod et al. \cite{mahbod19fusing}                 & Dermoscopy    & Ensemble              & ImageNet      & Selective     & Melanoma     \\
            Mahbod et al. \cite{mahbod20transfer}               & Dermoscopy    & Ensemble              & ImageNet      & FT            & Melanoma  \\
            Mukhlif et al.  \cite{mukhlif23incorporating}       & Dermoscopy    & Multiple              & ImageNet/Multiple & Selective & Melanoma     \\
            Hasan et al. \cite{hasan22dermoexpert}              & Dermoscopy    & CNN                   & ImageNet      & FT            & Melanoma     \\
            Spolaor et al. \cite{spolaor23fine}                 & Dermoscopy    & VGG                   & ImageNet      & Selective     & Melanoma     \\
            Tajbakhsh et al. \cite{tajbakhsh16convolutional}    & Endoscopy     & AlexNet               & ImageNet      & LP-FT       & Colonic Polyps\\
            Kim et al. \cite{kim21new}                          & Endoscopy     & AlexNet               & ImageNet      & LP            & Colonic Polyps     \\
            Patrini et al. \cite{patrini20transfer}             & Endoscopy     & Multiple              & ImageNet      & LP            & Frame selection    \\
            Liu et al. \cite{liu20finetuning}                   & Endoscopy     & Multiple              & ImageNet      & LP            & Gastric cancer   \\
            Jaafari et al. \cite{jaafari21towards}              & Endoscopy     & Inception Resnet      & ImageNet      & LP + Selective   & Surgical tools     \\            
            \hline
        \end{tabular}}
\end{table}

\subsection{X-ray and CT-scan}
Computed Tomography (CT) scans and X-rays are widely used in medical diagnostics, utilizing electromagnetic radiation to generate images of internal structures of the body, including bones, lungs, and heart. X-ray images, typically presented in grayscale, are effective in differentiating between dense and soft tissues. Chest X-rays are particularly common for diagnosing lung diseases such as pneumonia, tuberculosis, and COVID-19, as they provide information on various conditions from a single image. However, these images often face challenges such as low resolution, high noise, and limited soft tissue details, which complicate tasks such as classification and segmentation. The development of large X-ray datasets \cite{rajpurkar17chexnet, irvin19chexpert, bustos20padchest} has facilitated advancements in automated image analysis. Nevertheless, these datasets often contain imbalances and require multi-label classification due to the presence of multiple pathologies. In response, several studies have explored fine-tuning pre-trained networks for chest X-ray classification. Kermany et al. \cite{kermany18identifying} used transfer learning with an Inception v3 architecture, originally trained on ImageNet, to distinguish normal from pneumonia cases in chest X-rays. They proposed retraining only the final softmax layer, with the convolutional layers remaining unchanged, which yielded high-level recognition performance. Similarly, Yadav et al. \cite{yadav19deep} employed modified VGG-16 and InceptionV3 architectures, pre-trained on ImageNet, examining various fine-tuning approaches, including modifications in the number of fine-tuned layers, classifier size, dropout, and learning rate adjustments. Their results highlighted that VGG-16, with only the last classifier layer retrained, provided the most effective performance. Liang et al. \cite{liang20transfer} proposed a deep learning framework that combines residual thinking and dilated convolution for the diagnosis and detection of childhood pneumonia. They improved the performance by pre-training the model with the ChestX-ray14 dataset and conducting a full fine-tuning process. Chouhan et al. \cite{chouhan20novel} pre-trained several CNN architectures on ImageNet for pneumonia detection, adopting a linear probing approach for model adaptation. They also proposed an ensemble model where the outputs of the evaluated architectures (AlexNet, DenseNet, Inception, ResNet, and GoogleNet) were combined through a majority voting system. Apostolopoulos et al. \cite{apostolopoulos20covid} trained various CNN architectures, including VGG-19, MobileNet v2, Inception, and Xception, on a combined COVID-19 and pneumonia images dataset. They adopted a selective transfer learning approach, empirically testing different combinations of layer blocks for training. The study found that VGG-19 showed high accuracy, while MobileNet v2 was noted for its specificity in the detection of COVID-19. In the context of CT scans, Maghdid et al. \cite{maghdid20diagnosing} developed a COVID-19 detection model using a pre-trained AlexNet architecture on ImageNet, combined with a target dataset of X-ray and CT-scan images. This model underwent a complete retraining of network weights, leading to improved accuracy in detection. Ahuja et al. \cite{ahuja21deep} evaluated COVID-19 detection models from CT scans using various pre-trained ResNet architectures, applying a full fine-tuning approach. Their findings indicated that the ResNet-18 network, despite having fewer layers, achieved notable performance.

\subsection{MRI}
Magnetic Resonance Imaging (MRI) is a technique that uses magnetic fields and radio waves to produce high-resolution images of internal body structures. It is especially useful for examining brain tumors, helping in the diagnosis and segmentation of different types of tumors, such as glioma, meningioma, and pituitary adenoma. Several works have investigated the use of fine-tuning pre-trained networks for brain tumor classification tasks. Basaia et al. \cite{basaia19automated} designed a CNN algorithm to distinguish between healthy individuals, patients with Alzheimer’s disease, and those with mild cognitive impairment (MCI) on the path to Alzheimer’s. They used a CNN architecture pre-trained with the Alzheimer’s Disease Neuroimaging Initiative (ADNI) dataset, and applied a full fine-tuning approach. This method involved retraining all network layers, achieving an accuracy of up to 75\%. Likewise, Oh et al. \cite{oh19classification} implemented a full fine-tuning strategy for binary classification tasks related to Alzheimer’s disease. Their methodology involved using initial weights from a convolutional AutoEncoder (CAE) based unsupervised learning model, followed by fine-tuning with the target dataset, improving both performance and efficiency. Eitel et al. \cite{eitel19uncovering} proposed a transparent deep learning framework using 3D convolutional neural networks and layer-wise relevance propagation for diagnosing multiple sclerosis with MRI scans. The model was trained using four strategies: training from scratch, training only on the target dataset, training only with the ADNI dataset, and full fine-tuning on the target dataset with a pre-trained model using the ADNI dataset. The results indicated that fine-tuning greatly enhanced the identification of relevant regions. Jonsson et al. \cite{jonsson19brain} presented a 3D CNN-based model to predict brain age from T1-weighted MRIs. They used a ResNet architecture, initially freezing the convolutional layers so that only the final fully connected layers were trainable. Afterward, the model underwent retraining with data from a new scanning site, reducing the number of parameters to train and effectively increasing the prediction accuracy for new sites. Naser et al. \cite{naser20brain} integrated a CNN using U-net architecture for tumor analysis, specifically targeting tumor segmentation and grading. The study pre-trained a VGG-16 convolutional base for tumor segmentation, followed by retraining only the fully connected classifier for tumor grading. This approach achieved tumor grading accuracy at both image and patient levels of 0.89 and 0.9, respectively.

\subsection{Microscopy}
The availability of high-resolution gigapixel whole-slide images (WSI) has improved digital pathology and microscopy, enabling detailed and accurate cancer diagnosis through color images of tissues like skin, breast, and colon for identifying and grading various cancers including melanoma, breast cancer, and colon cancer. Several studies have taken advantage of fine-tuning pre-trained networks for this task. Nawaz et al. \cite{nawaz18classification} modified an AlexNet architecture to classify breast cancer histology images, replacing its original classification layers with a single convolutional layer and fine-tuning the last three layers using the ICIAR2018 dataset after pre-training on ImageNet. They achieved an image-wise classification accuracy of 81.25\% and a patch-wise accuracy of 75.73\%. Ferreira et al. \cite{ferreira18classification} utilized an Inception-ResNet-v2 architecture with pre-trained weights from ImageNet, initially retraining only its fully connected layers and later fine-tuning additional top layers with H\&E stained breast histology images. Bayramoglu \cite{bayramoglu16transfer} evaluated the performance of fine-tuning pre-trained models, including CNN architectures such as AlexNet and VGG-16, versus training from scratch for cell nuclei classification. Their fine-tuning strategy consisted of choosing different learning rates for pre-trained layers and added classification layers. They found that the fine-tuned models showed improved results and efficiency. Vesal et al. \cite{vesal18classification} applied full fine-tuning to Inception v3 and ResNet-50 architectures, initially trained on ImageNet, to classify breast histology images into sub-types. The fine-tuned ResNet-50 achieved a remarkable accuracy of 97. 5\%. Vrbancic et al. \cite{vrbancic20transfer} proposed an adaptive transfer learning approach for osteosarcoma detection in H\&E stained images using a VGG-19 architecture. This method involved selectively fine-tuning network layers determined by differential evolution optimization, showing improved performance over other training techniques.

\subsection{Dermoscopy}
Dermoscopy images are commonly utilized in the analysis of various types of skin lesions, such as melanomas, aiding in their diagnosis and segmentation. However, it presents significant challenges for automated detection due to high intra-class variability and inter-class similarities \cite{wang23intraclass}, making it difficult to find a consistent pattern or distinguish between benign and malignant lesions. The ISIC Challenge, despite offering a vast repository of melanoma cases, highlights these difficulties, as models trained on one year's data often fail to perform well on subsequent years \cite{cassidy22analysis}. Many studies have leveraged pre-trained networks and fine-tuning techniques to enhance the adaptation and robustness of their models for skin lesion analysis tasks. For example, Mahbod et al. \cite{mahbod19fusing} introduced a fully automated computerized method for classifying skin lesions from dermoscopic images using an ensemble of CNNs with different architectures that capture varying levels of feature abstraction. Each CNN architecture, such as AlexNet, VGGNet, and ResNet variations, underwent fine-tuning by freezing their initial layers and training the remaining layers using the ISIC 2017 skin lesion classification challenge dataset. This fine-tuning process involved variations in normalization techniques and optimizers. The ensemble approach, which combined the average classification output of individual CNNs across all variations, enhanced the performance compared to using single architectures alone. In a subsequent study \cite{mahbod20transfer}, the authors utilized the ISIC 2018 challenge test dataset and demonstrated that image cropping outperformed image resizing in a three-level fusion scheme of fine-tuned CNN architectures. They implemented a full fine-tuning strategy with different hyperparameter settings, applying a low learning rate and L2 regularization to the CNN layers, while the fully connected layers used a higher learning rate. Mukhlif et al. \cite{mukhlif23incorporating} proposed a Dual Transfer Learning approach, involving a two-phase fine-tuning process. Initially, a pre-trained model on ImageNet was fine-tuned on a source dataset containing unlabeled images from datasets similar to the target domain, with the initial layers frozen. In the second phase, intermediate layers retrained on the source dataset were also frozen, and only the remaining layers were fine-tuned on the target dataset. The authors evaluated this approach using four pre-trained models (VGG-16, Xception, ResNet-50, MobileNetV2) with the ISIC2020 skin cancer dataset, showing performance improvements across all architectures, with Xception achieving the highest performance. Hasan et al. \cite{hasan22dermoexpert} introduced DermoExpert, an automated skin lesion classification framework based on a hybrid CNN. In this framework, a batch of input images is simultaneously processed through three different Feature Map Generators (FMG), each pre-trained on ImageNet and fully fine-tuned on the target datasets. Evaluation results demonstrated that this approach achieved state-of-the-art performance on the ISIC 2016, 2017, and 2018 datasets. Spolaor et al. \cite{spolaor23fine} proposed various fine-tuning configurations for VGG architectures pre-trained on ImageNet for dermoscopic image classification. Their fine-tuning strategies involved selecting different architectures, freezing specific layers, and adjusting learning rates. The findings indicated that fine-tuning a VGG-16 architecture by freezing the top layers and fine-tuning the remaining layers resulted in the best performance for skin image classification.

\subsection{Endoscopy}
Endoscopic images are obtained with the use of an endoscope to produce images of internal organs of the body, such as the stomach, colon, and bladder. Endoscopy images are typically colorful, with high resolution and contrast between organ structures, such as mucosa, polyps, and ulcers. Tajbakhsh et al. \cite{tajbakhsh16convolutional} evaluated the impact of fine-tuning on four different medical imaging applications that involve classification, detection, and segmentation. They followed a systematic fine-tuning approach by first fine-tuning the last classification layer (linear probing) and then fine-tuning all layers of the network. They validated their proposed approach in a polyp detection task using images extracted from colonoscopy videos. They used an AlexNet architecture pre-trained on ImageNet. Their results showed that fine-tuning led to superior performance and accelerated the speed of convergence. Furthermore, their experiments showed that with the use of reduced training data, fine-tuned models dramatically surpassed the performance of conventional CNNs trained from scratch. Kim et al. \cite{kim21new} also focused on polyp detection from colonoscopy images using an AlexNet architecture but evaluated multiple sources for pre-training. Their results showed that linear probing fine-tuning on a model pre-trained on ImageNet generated the best accuracy performance. Patrini et al. \cite{patrini20transfer} proposed a strategy for the automatic selection of informative laryngoscopic video frames to reduce the amount of data to process for diagnosis. They followed a linear probing fine-tuning approach, freezing the CNN layers and only retraining the classifier layers. They evaluated six different CNN architectures pre-trained on ImageNet, with VGG-16 exhibiting the best performance compared to training from scratch for recognizing informative frames. Liu et al. \cite{liu20finetuning} proposed a transfer learning framework to recognize gastric lesion characteristics from gastric M-NBI images. Their results showed that linear-probing fine-tuning in a ResNet-50 architecture pre-trained on ImageNet achieved high accuracy results and convergence speed, while VGG-16 achieved the worst performance. Jaafari et al. \cite{jaafari21towards} classified endoscopic images to recognize surgical tools presence. They employed an Inception Resnet v2 architecture and the Cholec80 dataset. Their fine-tuning strategy consisted of first training only the last classification layer (linear probing) and then fine-tuning some of the CNN layers with a smaller learning rate. Their results showed higher classification performance compared to other state-of-the-art approaches.

While prior research has demonstrated the efficacy of fine-tuning pre-trained models to improve classification accuracy in medical imaging, there remains a notable gap in comparative analysis across different fine-tuning strategies and architectures, particularly when applied to diverse medical image modalities. This study seeks to bridge this gap by conducting a systematic comparison of various fine-tuning methodologies across a range of medical imaging domains and architectures.

\section{Materials and Methods}
\label{sec:3}

\subsection{Fine-tuning methods}

\begin{figure}[H]
	\centering
	\includegraphics[width=0.55 \textwidth]{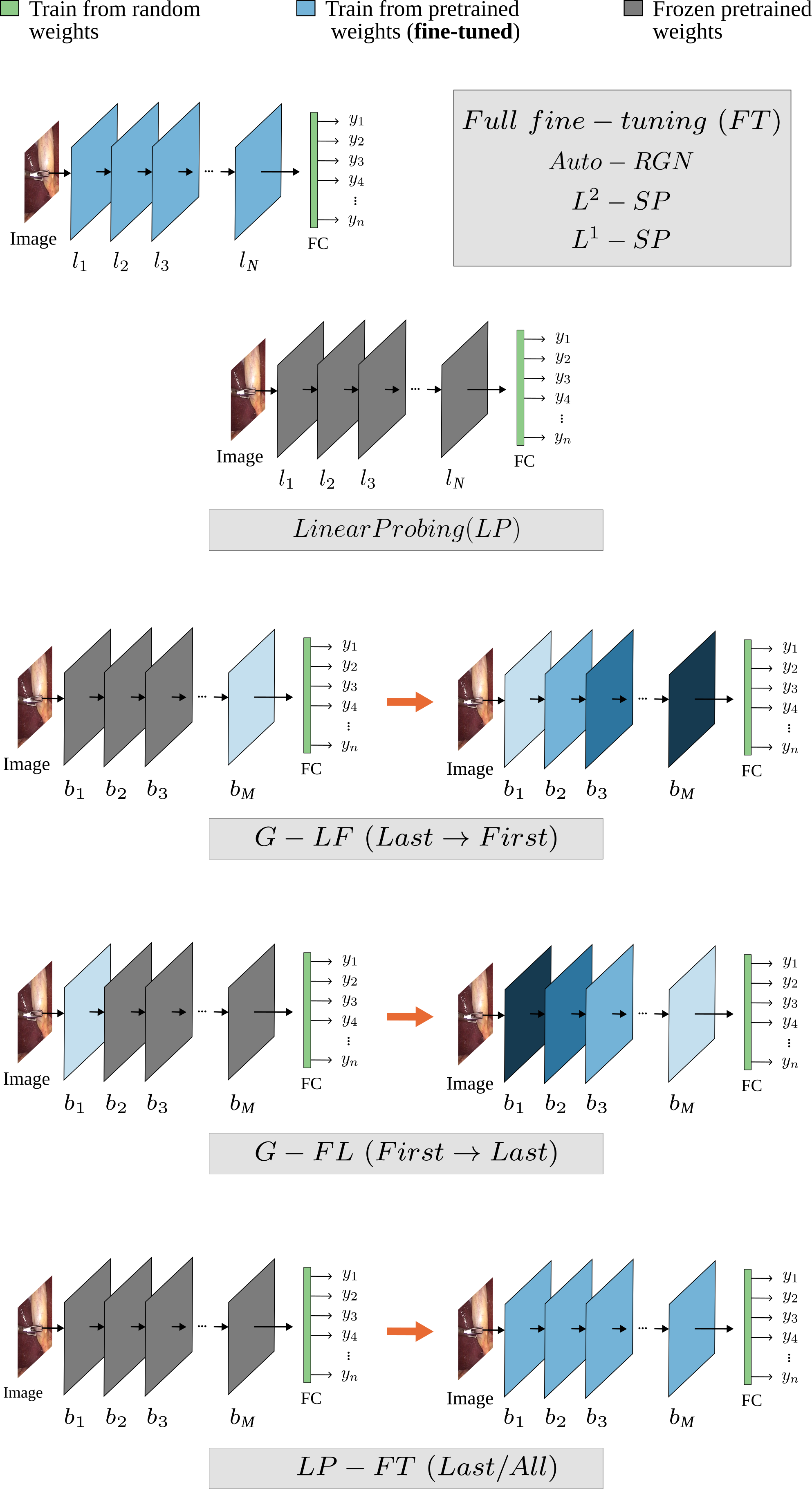}
	\caption{Illustration of evaluated fine-tuning methods. The blue blocks represent fine-tunable layers, while gray blocks are frozen layers (non fine-tunable). From top to bottom: (1) Fine-tuning strategies where all layers $\{l_1, \dots, l_N, FC\}$ are fine-tuned; (2) Linear Probing, where only the classifier layers $FC$ are retrained; (3) Gradual Unfreezing fine-tunable layers, starting from the last layer; (4) Gradual Unfreezing fine-tunable layers, starting from the first layer; (5) Training only the classifier layer initially, followed by fine-tuning all layers.}
	\label{fig:5}
\end{figure}

In this study, we explore eight different fine-tuning strategies for adapting pre-trained models to medical image datasets, as illustrated in Fig.\ref{fig:5}. This examination extends beyond the traditional fine-tuning methods discussed in Section \ref{sec:2}, incorporating recent strategies that have shown promising results in general image classification tasks. The fine-tuning strategies that we investigate include:

\subsubsection{Full Fine-tuning (FT)}
 Full Fine-tuning is a standard and widely-employed method for transfer learning that involves training all layers of the pre-trained model alongside the newly added classifier layer on the target dataset. While this method thoroughly adapts the model to the new task, it may lead to overfitting or a loss of original features \cite{lee23surgical}.

\subsubsection{Linear Probing (LP) \cite{kumar22fine}}
In this approach, we freeze all layers of the pre-trained model except the last one (layer $l_N$), which is the randomly initialized classifier layer, and train it on the target dataset. This method preserves the original features of the pre-trained model but might not fully capture the nuances of the new task.

\subsubsection{Gradual Unfreeze (Last → First) (G-LF) \cite{howard18universal}}
We progressively unfreeze blocks of layers ($b_i$) in the pre-trained model, starting from the last one ($b_M$) and moving to the first ($b_1$), training them alongside the new classifier layer on the target dataset. This controlled approach adapts the model to the new task, beginning with the most task-specific layers and progressing to the most general ones.

\subsubsection{Gradual Unfreeze (First → Last) (G-FL) \cite{mukherjee20distilling}}
This method is the reverse of G-LF. We gradually unfreeze each block of layers ($b_i$), starting from the first ($b_1$) and progressing to the last ($b_M$), training them with the new classifier layer. This strategy adapts the pre-trained model to the new task by starting from the most general layers and moving towards the more task-specific ones.

\subsubsection{Gradual Unfreeze (Last/All) (LP-FT)\cite{kumar22fine}}
Combining Linear Probing and Full Fine-tuning, we initially train only the last layer of the pre-trained model ($l_N$) while freezing the remaining layers. After a set number of epochs, we unfreeze all layers and continue training with the new classifier. This balanced approach initially preserves the original features, then updates them according to the new task requirements.

\subsubsection{$L^1$-\textit{SP} Regularization \cite{li18explicit}}
We incorporate an $L^1$-norm based sparse regularization term into the objective function of the pre-trained model and train it with the new classifier layer on the target dataset. This method encourages sparsity in layer weights, helping to prevent overfitting and retain useful features from the source task. The regularization term is defined as

\begin{equation}
    \label{eq:1}
    \gamma(\omega) = \alpha {|| \omega_{s} - \omega^{0}_{s} ||}_{1} + \beta {|| \omega_{\hat{s}} ||}_{1} ,
\end{equation}

where $\omega_s$ represents the model's parameter vector fine-tuned on the target task, $\omega^{0}$ is the parameter vector from the source task, serving as the starting point (\textit{-SP}) for fine-tuning, $\omega_{\hat{s}}$ is the parameter vector for the last classifier layer, and $\alpha$ and $\beta$ are arbitrary positive coefficients. The regularized objective function, $\hat{J}(\omega)$, combines the standard objective function ${J}(\omega)$ with the regularizer $\gamma(\omega)$:

\begin{equation}
    \label{eq:2}
     \hat{J}(\omega) = J(\omega) + \gamma(\omega) .
\end{equation}

\subsubsection{$L^2$-\textit{SP} Regularization \cite{li18explicit}}
Similar to $L^1$-\textit{SP} Regularization, this method applies an $L^2$-norm based regularization to each layer of the pre-trained model, training them with the new classifier layer. This approach, while similar to $L^1$-\textit{SP}, penalizes larger weights more heavily than smaller ones:

\begin{equation}
\label{eq:3}
\gamma(\omega) = \alpha || \omega_{s} - \omega^{0}{s} ||^2_{2} + \beta || \omega_{\hat{s}} ||^2_{2}
\end{equation}

\subsubsection{Auto-RGN \cite{lee23surgical}}
This method involves fine-tuning the pre-trained model with the new classifier layer on the target dataset, adjusting the learning rate for each layer based on its relative gradient norm (RGN). The RGN is calculated by dividing the gradient norm $||g_{i}||$ by the parameter vector norm $||\omega_{i}||$ for each layer $l_i$:

\begin{equation}
\label{eq:4}
RGN(l_{i}) = \frac{{||g_{i}||_2}}{||\omega{i}||_2} .
\end{equation}

The learning rate for each layer is then modified in proportion to its normalized RGN, allocating higher learning rates to more informative layers and lower rates to less informative ones. This method fine-tunes the model by focusing on the most relevant features for the new task.

 \subsection{Metrics}
We employ various evaluation metrics, adhering to the recommended metrics for each dataset in their original publications. The metrics reported are:
 
 \begin{itemize}
     \item Mean Average Precision (mAP): This metric calculates the average precision across all classes.
         \begin{equation}
             mAP = \frac{1}{C} \sum_{j=1}^{C} \text{AP}_j
         \end{equation}
         Where \(C\) is the number of classes, and \(\text{AP}_j\) is the Average Precision for class \(j\).
         
     \item Macro-Accuracy (Acc): This measures the accuracy across all classes. 
         \begin{equation}
             Acc = \frac{1}{C} \sum_{j=1}^{C} \frac{TP_j + TN_j}{TP_j + TN_j + FP_j + FN_j}
         \end{equation}
        where \(C\) is the number of classes, \(TP_j\) denotes the true positive count for class \(j\), \(TN_j\) the true negative count, \(FP_j\) the false positive count, and \(FN_j\) the false negative count.
                 
     \item Area under receiving operating characteristic (AUROC): This metric assesses the model's capability to distinguish between positive and negative examples.
         \begin{equation}
             AUROC = \frac{1}{C} \sum_{j=1}^{C} AUROC_j
         \end{equation}
         Where \(C\) is the number of classes, and \(AUROC_j\) is the Area Under the ROC Curve for class \(j\).
         
     \item Cohen Kappa ($\kappa$): This metric evaluates the agreement between the model predictions and the ground truth, considering the probability of chance agreement.
         \begin{equation}
             \kappa = \frac{p_o - p_c}{1 - p_c}
         \end{equation}
         Where \(p_o\) is the observed agreement and \(p_c\) is the expected agreement by chance.
         
     \item Macro F1 score: This calculates the harmonic mean of precision and recall for all classes.
         \begin{equation}
             F1 = \frac{1}{C} \sum_{j=1}^{C} \frac{2 \cdot precision_j \cdot recall_j}{precision_j + recall_j}
         \end{equation}
       where precision is the ratio of true positives to the total of true positives and false positives, and recall is the ratio of true positives to the total of true positives and false negatives.
 \end{itemize}
 
\subsection{Datasets}
Our experiments are conducted on six publicly available datasets from five distinct medical domains: X-ray, MRI, Histology, Dermoscopy, and Endoscopic Surgery. These datasets encompass a wide range of tasks and present various challenges, including class imbalance, data scarcity, label uncertainty, and inter-patient variability. A detailed description of each dataset is provided below.

\begin{figure}[H]
	\centering
	\includegraphics[width=0.55 \textwidth]{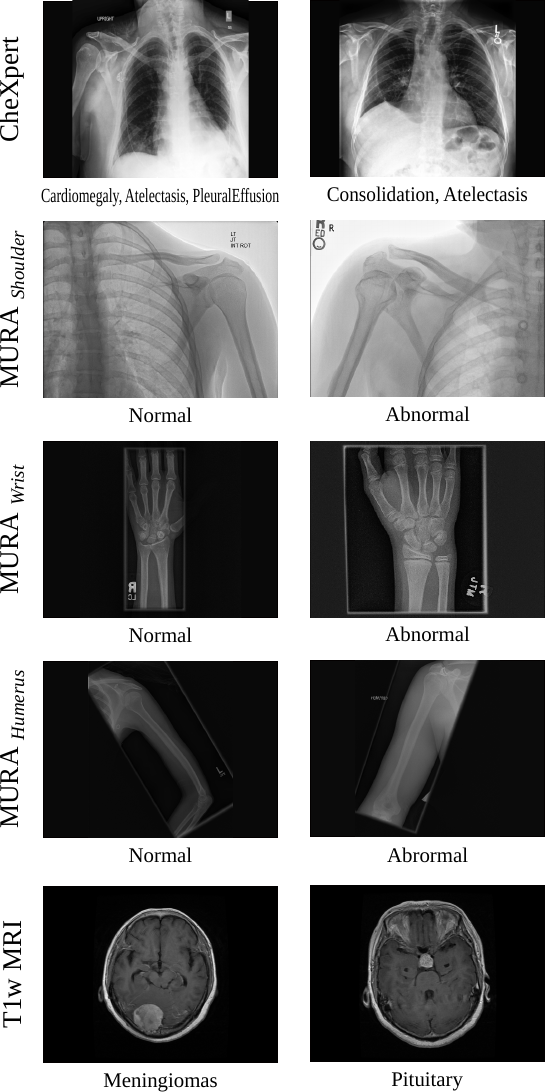}
	\caption{Representative images from the CheXpert, MURA, and T1w MRI datasets, showcasing a variety of radiographic and MRI imaging types.}
	\label{fig:2}
\end{figure}

\begin{figure}[H]
	\centering
	\includegraphics[width=0.60 \textwidth]{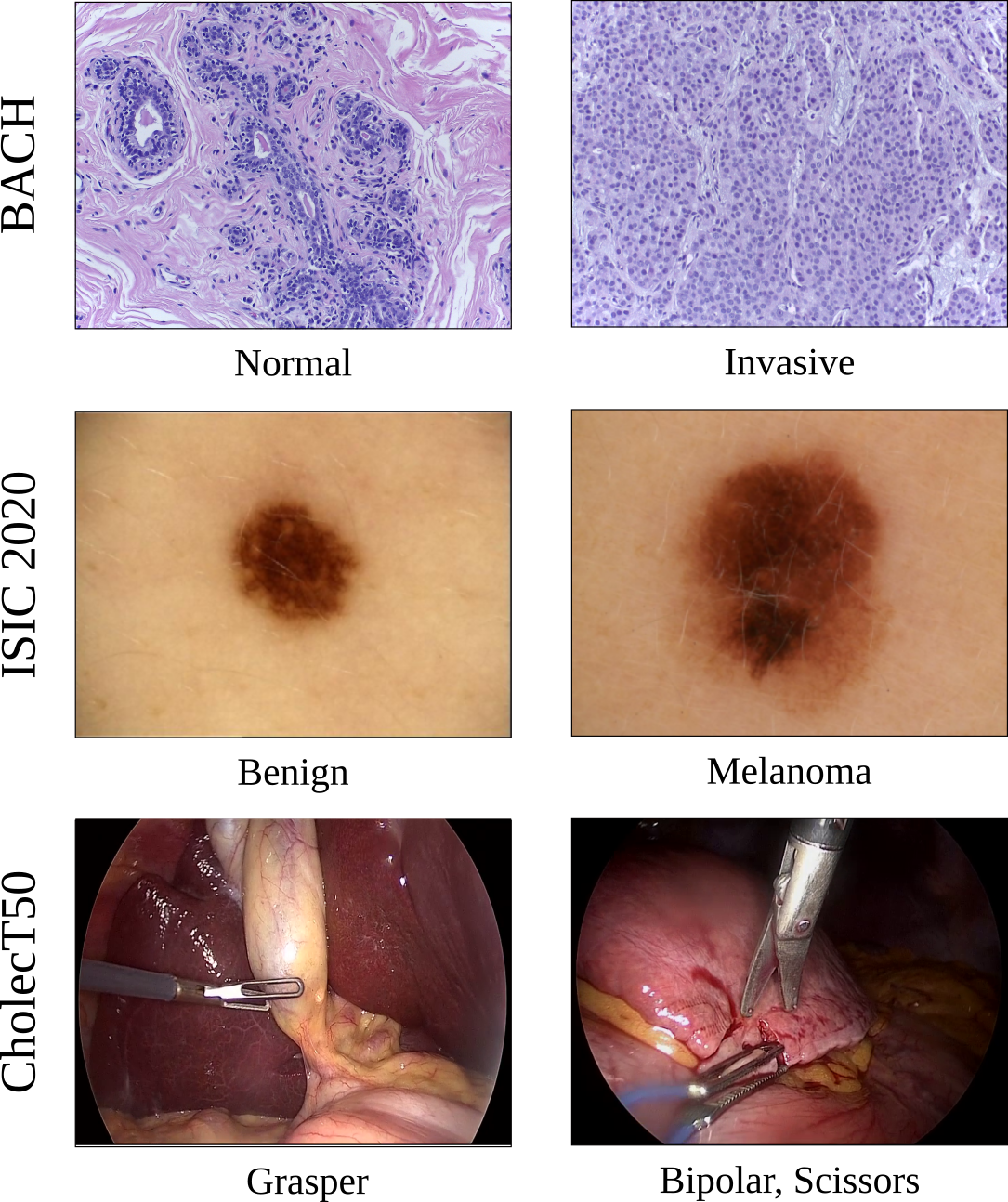}
	\caption{Sample images from the BACH, ISIC 2020, and CholecT50 datasets, illustrating histology, dermoscopy, and endoscopic surgery imaging modalities.}
	\label{fig:3}
\end{figure}

\begin{itemize}
 
    \item CheXpert \cite{irvin19chexpert}: This dataset includes 224,316 chest radiographs from 65,240 patients at Stanford Hospital, labeled for 14 common chest radiographic observations. The grayscale images have been resized to 320 pixels, maintaining their aspect ratio. The dataset is split into training, validation, and test sets, with radiographs from 64,540, 200, and 500 patients, respectively. Each set comprises images from different patients, with the training set containing one image per patient and the validation and test sets featuring multiple images per patient. Our focus is on five disease classes: cardiomegaly, edema, consolidation, atelectasis, and pleural effusion, as recommended in \cite{yuan21large}. This dataset is characterized by imbalances in the number of images and labels for each class and patient. Example images are presented in Fig.~\ref{fig:2}.

    \item MURA \cite{rajpurkar17mura}: Consisting of 40,561 multi-view radiographs of the upper extremity from 12,173 patients, this dataset covers the shoulder, humerus, elbow, forearm, wrist, hand, and finger. These images vary in resolution and aspect ratio and are categorized as normal or abnormal based on the presence of abnormalities. For our experiment, we focus on the categories \textit{Shoulder}, \textit{Wrist}, and \textit{Humerus}. The dataset presents a class imbalance within these categories, for \textit{Shoulder} category, there are 1,507 normal and 1,260 abnormal studies; for \textit{Wrist}, there are 1,396 normal and 2,078 abnormal studies; and the \textit{Humerus} category comprises 336 normal and 383 abnormal studies.
    Example images are shown in Fig.~\ref{fig:2}.
    
    \item T1w MRI \cite{chen15enhanced, chen15dataset}: This dataset contains 3,064 contrast-enhanced magnetic resonance images (CE-MRI) of brain tumors from 233 patients, collected between 2005 and 2010 in two hospitals in China. The images include three types of brain tumors: meningiomas, gliomas, and pituitary tumors, captured from axial, coronal, and sagittal views. The dataset is imbalanced, with respective counts of 708 meningiomas, 1,426 gliomas, and 930 pituitary tumors. Example images are shown in Fig.~\ref{fig:2}.

    \item BACH \cite{aresta19bach}: Comprising 400 high-resolution images of breast histology microscopy slides stained with hematoxylin and eosin (H\&E), this dataset was acquired from three different hospitals in Portugal. The 2048x1536 pixel images are classified into four categories based on breast cancer presence and type: normal, benign, in situ carcinoma, and invasive carcinoma. The dataset is balanced, with 100 images per class. Example images are shown in Fig.~\ref{fig:3}.

    \item ISIC 2020 \cite{rotemberg21patient}: This dataset includes 33,126 dermoscopy images collected from over 2,000 patients with various skin lesions. After excluding 425 duplicate images, the training set consists of 32,701 unique images. The images, which range in resolution from 640x480 to 6000x4000 pixels, are classified as benign or malignant based on the presence of melanoma. The dataset is significantly imbalanced, with only 584 (1.8\%) images confirmed histopathologically as melanoma. Example images are shown in Fig.~\ref{fig:3}.
    
    \item CholecT50 \cite{nwoye23cholectriplet}: This dataset includes 50 video recordings of laparoscopic cholecystectomy surgeries, annotated at 1 frame per second (fps) with binary presence of action triplets in the format $<$instrument, verb, target$>$. The triplets comprise 6 instruments, 10 verbs, and 15 targets, totaling 100 triplet classes. For our experiments, we focus on the \textit{Instrument} labels for each component of the triplet. The videos were downsampled to 6 frames per minute (fpm), yielding a total of 10,109 frames. Example images are shown in Fig.~\ref{fig:3}.

\end{itemize}

\begin{table}[H]
\centering
\begin{threeparttable}
    \caption{Distribution of samples in experimental datasets}\label{tab:2}
        \begin{tabular}{l c c c}
            \hline
            \bf{Dataset}    	       &\bf{Training}  &\bf{Evaluation}	&\bf{Classes} \\
            \hline            
            CheXpert                    & 64540         & 200            & 5\textsuperscript{a}\\
            MURA-\it{Shoulder}          & 2694          & 173            & 2 \\
            MURA-\it{Wrist}             & 3267          & 207            & 2 \\
            MURA-\it{Humerus}           & 587           & 132            & 2 \\
            T1w MRI                     & 2522          & 542            & 3 \\
            BACH            	        & 360           & 40             & 4\\  
            ISIC 2020                   & 26160         & 6541           & 2\\        
            CholecT50-\it{Instrument}   & 7980          & 2129           & 6\textsuperscript{a}\\ 
            \hline
        \end{tabular}
    \footnotetext[1]{Footnote}
    \begin{tablenotes}
      \item[a]{Multilabel}
    \end{tablenotes}
\end{threeparttable}
\end{table}

\subsection{Image preprocessing}
We resize the input images to maintain their aspect ratio while adjusting the shorter side to 232, 256, and 256 pixels for the ResNet-50, DenseNet-121, and VGG-19 models, respectively. Subsequently, we normalize the images using the mean and standard deviation values from ImageNet. The final step involves cropping the images to a uniform size of 224×224 pixels. To ensure consistent results and to specifically assess the impact of the fine-tuning strategies, we refrain from applying any data augmentation techniques. However, we recognize that employing higher resolution images or data augmentation methods might enhance performance, as indicated in prior research \cite{cui18large, cubuk19autoaugment}.

\subsection{Pre-trained model architectures}

\begin{figure}[H]
	\centering
	\includegraphics[width=1.0 \textwidth]{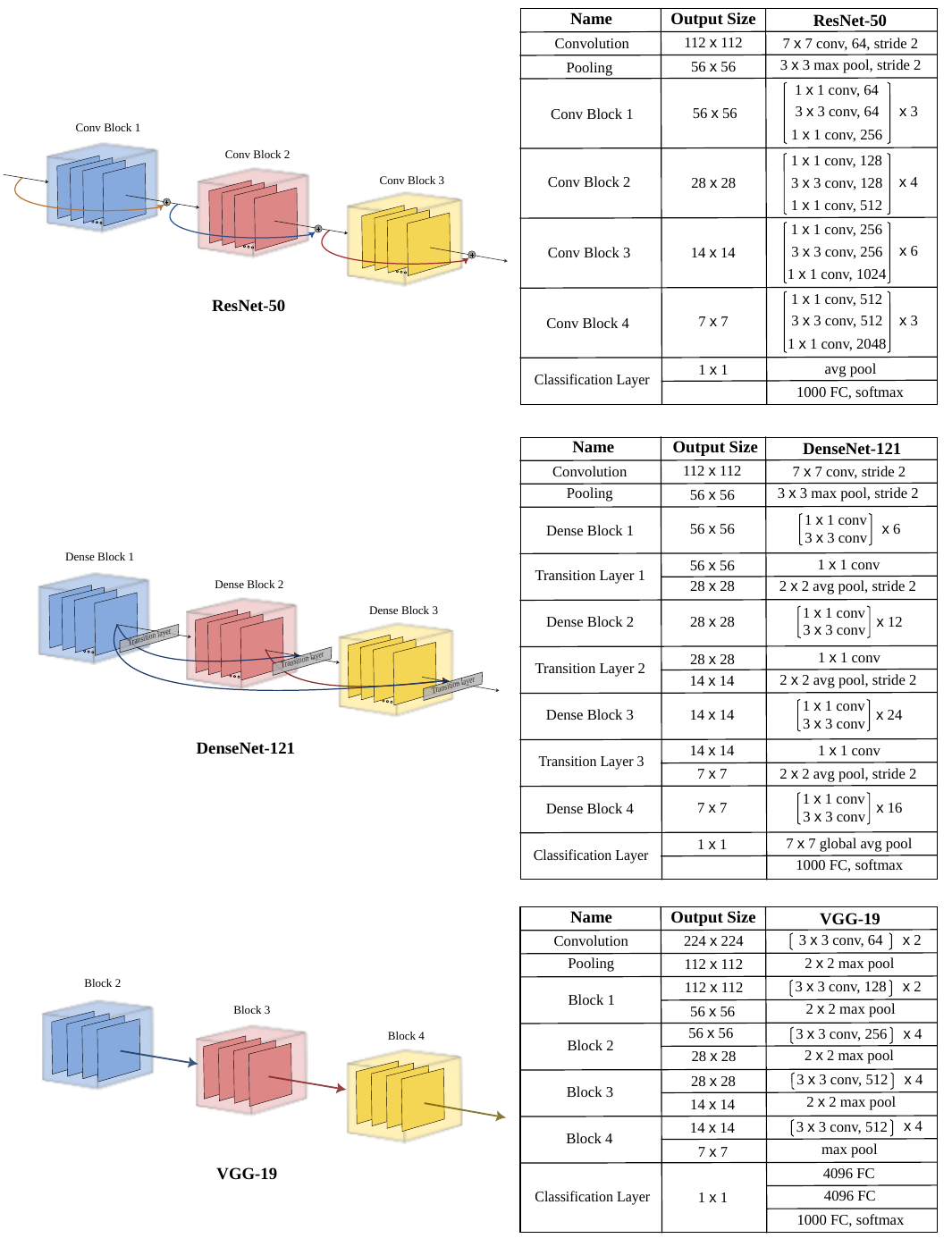}
	\caption{Architectures of the pre-trained models.}
	\label{fig:4}
\end{figure}

We utilize three distinct CNN architectures, as depicted in Fig.~\ref{fig:4}, which not only provide different mechanisms of learning and feature extraction but also vary in their depth and complexity. This variety ensures that our analysis of fine-tuning strategies encompasses a broad range of scenarios, offering a more generalizable understanding of their effects. Each architecture is pre-trained on the ImageNet dataset.

\begin{itemize}
\item ResNet-50 \cite{he16deep}: This 50-layer CNN features shortcut connections that bypass some layers, effectively preventing the vanishing gradient problem. It comprises 25.6 million parameters. For the implementation of the gradual unfreezing fine-tuning methods, the parameters are organized into six parameter blocks.
\item DenseNet-121 \cite{huang17densely}: A 121-layer CNN, DenseNet-121 incorporates dense connections that connect each layer to every other layer in a sequential fashion. This design enhances feature propagation and reuse while minimizing the number of parameters. It contains 8 million parameters. For the implementation of some of the gradual unfreezing fine-tuning methods, the parameters are organized into nine parameter blocks.
\item VGG-19 \cite{simonyan15deep}: Comprising 19 layers, VGG-19 uses small (3x3) convolution filters with an increasing number of filters in each subsequent layer. It has 143.7 million parameters. For the implementation of some of the gradual unfreezing fine-tuning methods, the parameters are organized into six parameter blocks.
\end{itemize}

In each model, we replace the final fully connected layer with a new layer that is randomly initialized and appropriately sized for the target dataset.

\subsection{Parameter Settings}
In each experiment, we conducted three trials using different random seeds and reported the mean and standard error of the metrics. The learning rate was adjusted among three values: 1e-3, 1e-4, and 1e-5. We employed the Adam optimizer and trained the models for 50 epochs, incorporating early stopping based on a patience interval of 5 epochs. The batch size was set at 64 for all experiments.

\section{Results}
\label{sec:4}

In this section, we present the results of our experiments on eight medical image classification datasets, employing three different pre-trained models (ResNet-50, DenseNet-121, and VGG-19) and eight distinct fine-tuning methods comprising five medical domains: X-ray, MRI, histology, dermoscopy and endoscopic. The mean and standard deviation of the metrics are reported over three runs with different random seeds.

\subsection{X-ray}
\begin{figure}[H]
	\centering
	\includegraphics[width=0.9\textwidth]{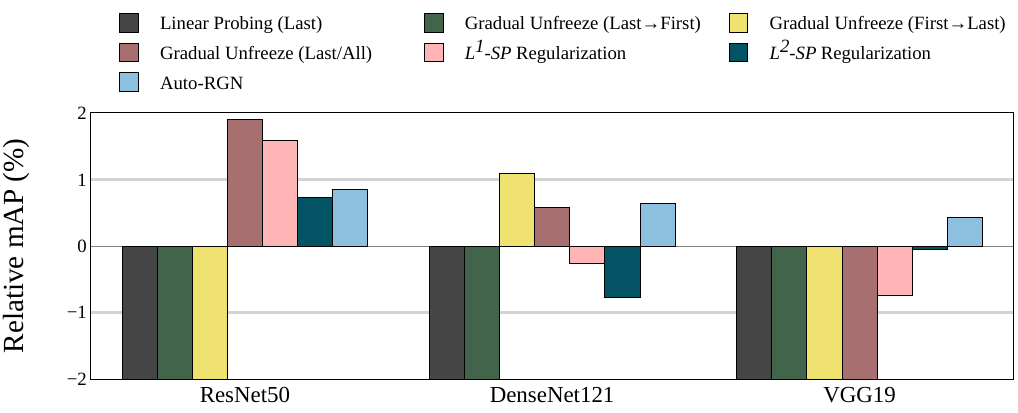}
	\caption{Relative mAP for CheXpert, which is calculated as the percentage change in mAP compared to Full fine-tuning (\textit{FT}).}
	\label{fig:6}
\end{figure}

\begin{figure}[H]
	\centering
	\includegraphics[width=0.9\textwidth]{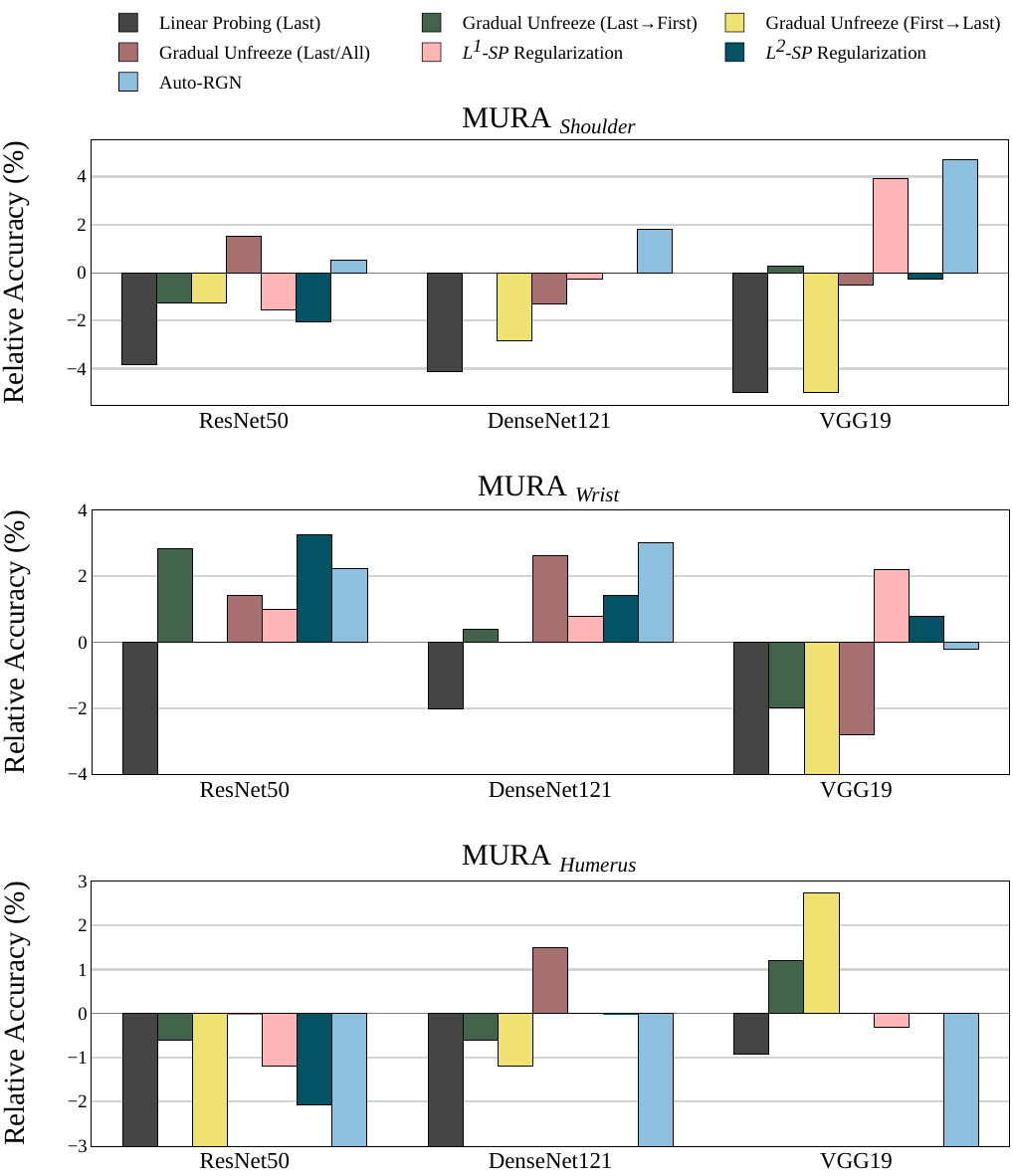}
	\caption{Relative Accuracy for MURA-\textit{Shoulder}, MURA-\textit{Wrist} and  MURA-\textit{Humerus}, which is calculated as the percentage change in Accuracy compared to Full fine-tuning (\textit{FT}).}
	\label{fig:7}
\end{figure}

For the CheXpert dataset, we used two evaluation metrics: mean Average Precision (mAP) and Area Under the Receiver Operating Characteristic Curve (AUROC). The results for CheXpert are detailed in Table~\ref{tab:3}, and the relative mAP improvement over conventional Full Fine-tuning (\textit{FT}) is depicted in Fig.\ref{fig:6}. For the MURA dataset, we evaluate three subsets representing different target tasks (MURA-\textit{Shoulder}, MURA-\textit{Wrist}, and MURA-\textit{Humerus}) using accuracy and Kappa metrics. The results for these subsets are reported in Tables \ref{tab:4}, \ref{tab:5}, and \ref{tab:6}, respectively. The relative Accuracy improvement over conventional \textit{FT} for the three MURA tasks is illustrated in Fig.~\ref{fig:7}. 

\begin{table}[H]
\newrobustcmd{\B}{\bfseries}
\newrobustcmd{\U}{\underline}
 \centering
\caption{Comparison of accuracy and Kappa along with standard errors across three runs for various fine-tuning methods, evaluated on the CheXpert dataset. The top scores for each metric and architecture are highlighted in bold, and the overall top score across all architectures is underlined.}
\scalebox{0.9}{
\begin{tabular}{l  c c  c c  c c}
\cline{1-7}
                	& \multicolumn{2}{c}{\bf{ResNet-50}}  	& \multicolumn{2}{c}{\bf{DenseNet-121}} & \multicolumn{2}{c}{\bf{VGG-19}}    \\ 
\bf{Method}                         	         & mAP      	       & AUROC   	& mAP   & AUROC	& mAP   & AUROC   \\ \hline
Full Fine-tuning (\textit{FT})                   & 71.98 (0.7)	     & 87.72 (0.0)	& 72.17 (0.7)	& 87.10 (0.4)	& 72.41 (0.3)	& 87.00 (0.0)	\\
Linear Probing (\textit{LP})                     & 60.61 (0.2)	     & 79.93 (0.0)	& 59.27 (0.2)	& 81.35 (0.1)	& 64.36 (0.5)	& 82.49 (0.4)	\\
Grad. Last $\rightarrow$ First (\textit{G-LF})   & 69.96 (1.2)	     & 86.04 (0.4)	& 68.84 (0.4)	& 85.72 (0.4)	& 67.00 (0.6)	& 83.78 (0.3)	\\
Grad. First $\rightarrow$ Last (\textit{G-FL})   & 70.35 (0.5)	     & 86.00 (0.2)	&\B72.96 (0.3)	& 87.07 (0.1)	& 65.48 (0.6)	& 82.63 (0.2)	\\
Grad. Last/All (\textit{LP-FT})                  &\B{\U{73.35} (0.6)} & 87.51 (0.2)	& 72.59 (0.5)	& 87.10 (0.1)	& 70.70 (0.8)	& 86.51 (0.4)	\\
$L^1$-$SP$ Regularization                   	 & 73.12 (0.4)	      & \B{\U{87.85} (0.2)}	& 71.98 (0.1)	& \B87.78 (0.1)	& 71.86 (0.4)	& 86.77 (0.2)	\\
$L^2$-$SP$ Regularization                   	 & 72.50 (0.5)	    & 87.73 (0.2)	& 71.61 (0.7)	& 86.54 (0.3)	& 72.37 (0.4)	& 86.98 (0.0)	\\
Auto-RGN                                    	 & 72.59 (0.2)	    & 87.99 (0.3	   &72.64 (0.5)	& 87.61 (0.1)	&\B72.72 (0.3)	& \B87.06 (0.1)	\\ \hline
\end{tabular}}
\label{tab:3}
\end{table}

We observe varying performance of the fine-tuning methods depending on the architecture, metric, and dataset. Among all evaluated fine-tuning methods, Linear Probing (\textit{LP}) consistently underperforms, indicating that only training the last layer of the pre-trained model alone is insufficient for this dataset. However, Gradual Unfreeze (Last/All) , which combines \textit{LP} with a subsequent \textit{FT} step (\textit{LP-FT}) appears effective when paired with ResNet-50 and DenseNet-121, showing consistent improvement across all experiments. Yet, when combined with VGG-19, no significant effect or even a decrease in performance is observed. Other gradual unfreezing methods (\textit{G-LF} and \textit{G-FL}) generally do not offer substantial benefits, except in MURA-\textit{Humerus} where \textit{G-FL} with VGG-19 achieved the best performance. The regularization techniques ($L^1$-\textit{SP} and $L^2$-\textit{SP}) enhance performance on all architectures for the MURA-\textit{Wrist} dataset. In other datasets, $L^2$-\textit{SP} shows limited improvement, while $L^1$-\textit{SP} boosts the performance of ResNet-50 in CheXpert and VGG-19 in MURA-\textit{Shoulder}. Adjusting the learning rate based on the relative gradient norm (Auto-RGN) is effective in the X-ray domain, outperforming \textit{FT} in all architectures, metrics, and X-ray datasets evaluated, except for MURA-\textit{Humerus}.

Regarding the three architectures, no single model consistently excels in all experiments. ResNet-50 yields better results for CheXpert, while VGG-19 is more effective in MURA-\textit{Shoulder} and MURA-\textit{Wrist}, and DenseNet-121 performs best in MURA-\textit{Humerus}. The smaller sample size in MURA-\textit{Humerus} (around 500 samples) compared to MURA-\textit{Shoulder} and MURA-\textit{Wrist} poses a challenge for fine-tuning, with ResNet-50 only showing reduced performance across all fine-tuning strategies, DenseNet-121 improving solely with \textit{LP-FT}, and VGG-19 benefiting from the gradual unfreezing approaches (\textit{G-LF} and \textit{G-FL}). The performance difference between the architectures is more pronounced for MURA than for CheXpert, potentially due to the disparity in the number of classes (14 for CheXpert vs. 2 for MURA) and dataset size (CheXpert having over 60,000 samples compared to less than 3,000 per task for MURA). 

\begin{table}[H]
\newrobustcmd{\B}{\bfseries} 
\newrobustcmd{\U}{\underline}
\centering
\caption{Comparison of accuracy and Kappa along with standard errors across three runs for various fine-tuning methods, evaluated on the MURA-\textit{Shoulder} dataset. The top scores for each metric and architecture are highlighted in bold, and the overall top score across all architectures is underlined.}
\scalebox{0.9}{
\begin{tabular}{l  c c  c c  c c}
\cline{1-7}
                	& \multicolumn{2}{c}{\bf{ResNet-50}}  	& \multicolumn{2}{c}{\bf{DenseNet-121}} & \multicolumn{2}{c}{\bf{VGG-19}}    \\ 
\bf{Method}                            	& Accuracy      	& Kappa   	& Accuracy   & Kappa	& Accuracy   & Kappa \\ \hline
Full Fine-tuning (\textit{FT})            	& 75.33 (1.6)	& 50.64 (3.3)	& 75.14 (1.0)	& 50.22 (2.0)	& 73.60 (0.1)	& 47.15 (0.3)	\\
Linear Probing (\textit{LP})                     & 72.44 (0.5)	& 44.94 (1.1)	& 72.06 (1.1)	& 43.81 (2.4)	& 69.94 (0.6)	& 39.93 (1.2)	\\
Grad. Last $\rightarrow$ First (\textit{G-LF})   & 74.37 (0.3)	& 48.66 (0.7)	& 75.14 (0.3)	& 50.32 (0.7)	& 73.79 (1.0)	& 47.53 (2.0)	\\
Grad. First $\rightarrow$ Last (\textit{G-FL})   & 74.37 (1.8)	& 48.69 (3.6)	& 73.02 (2.1)	& 46.16 (4.3)	& 69.94 (1.5)	& 39.82 (3.1)	\\
Grad. Last/All (\textit{LP-FT})               	&\B76.48 (0.8)	& \B52.91 (1.7)	& 74.18 (0.5)	& 48.37 (1.1)	& 73.21 (1.0)	& 46.44 (2.0)	\\
$L^1$-$SP$ Regularization               	& 74.17 (1.0)	& 48.25 (2.0)	& 74.94 (1.1)	& 49.80 (2.3)	& 76.49 (1.1)	& 52.90 (2.3)	\\
$L^2$-$SP$ Regularization               	& 73.79 (0.7)	& 47.41 (1.6)	& 75.14 (1.8)	& 50.26 (3.7)	& 73.41 (0.0)	& 46.78 (0.0)	\\
Auto-RGN 	                            	&75.72 (0.6)	& 51.47 (1.2)	&\B76.49 (0.5)	& \B53.05 (1.0)	&\B{\U{77.07} (0.3)}	& \B{\U{54.09} (0.7)}	\\ \hline
\end{tabular}}
\label{tab:4}
\end{table}

\begin{table}[H]
\newrobustcmd{\B}{\bfseries}
\newrobustcmd{\U}{\underline}
\centering
\caption{Comparison of accuracy and Kappa along with standard errors across three runs for various fine-tuning methods, evaluated on the MURA-\textit{Wrist} dataset. The top scores for each metric and architecture are highlighted in bold, and the overall top score across all architectures is underlined.}
\scalebox{0.9}{
\begin{tabular}{l  c c  c c  c c}
\cline{1-7}
                	& \multicolumn{2}{c}{\bf{ResNet-50}}  	& \multicolumn{2}{c}{\bf{DenseNet-121}} & \multicolumn{2}{c}{\bf{VGG-19}}    \\ 
\bf{Method}                            	& Accuracy      	& Kappa   	& Accuracy   & Kappa	& Accuracy   & Kappa \\ \hline
Full Fine-tuning (\textit{FT})            	& 79.54 (0.4)	& 57.98 (0.9)	& 80.03 (1.0)	& 59.05 (2.3)	& 80.83 (0.8)	& 60.52 (1.6)	\\
Linear Probing (\textit{LP})                     & 75.52 (1.4)	& 50.00 (2.8)	& 78.42 (0.4)	& 55.67 (0.9)	& 75.84 (0.4)	& 50.45 (0.8)	\\
Grad. Last $\rightarrow$ First (\textit{G-LF})   & 81.80 (0.5)	& 62.63 (1.2)	& 80.35 (0.1)	& 59.72 (0.2)	& 79.22 (0.2)	& 57.19 (0.5)	\\
Grad. First $\rightarrow$ Last (\textit{G-FL})   & 79.55 (0.7)	& 58.01 (1.4)	& 80.03 (0.8)	& 59.33 (1.8)	& 76.81 (0.2)	& 51.94 (0.4)	\\
Grad. Last/All (\textit{LP-FT})               	& 80.67 (1.0)	& 60.28 (1.8)	& 82.12 (0.2)	& 63.34 (0.4)	& 78.58 (0.9)	& 55.78 (2.0)	\\
$L^1$-$SP$ Regularization               	&80.35 (0.4)	& 59.64 (0.9)	&80.67 (0.9)	& 60.29 (1.9)	&\B{\U{82.61} (0.8)}	& \B{\U{64.33} (1.6)}	\\
$L^2$-$SP$ Regularization               	&\B82.12 (1.1)	& \B63.40 (2.3)	& 81.16 (0.5)	& 61.25 (1.0)	& 81.48 (0.6)	& 62.03 (1.4)	\\
Auto-RGN 	                            	& 81.32 (0.4)	& 61.66 (0.8)	&\B82.44 (0.8)	& \B64.12 (1.7)	& 80.67 (0.7)	& 60.51 (1.4)	\\ \hline
\end{tabular}}
\label{tab:5}
\end{table}

\begin{table}[H]
\newrobustcmd{\B}{\bfseries}
\newrobustcmd{\U}{\underline}
\centering
\caption{Comparison of accuracy and Kappa along with standard errors across three runs for various fine-tuning methods, evaluated on the MURA-\textit{Humerus} dataset. The top scores for each metric and architecture are highlighted in bold, and the overall top score across all architectures is underlined.}
\scalebox{0.9}{
\begin{tabular}{l  c c  c c  c c}
\cline{1-7}
                	& \multicolumn{2}{c}{\bf{ResNet-50}}  	& \multicolumn{2}{c}{\bf{DenseNet-121}} & \multicolumn{2}{c}{\bf{VGG-19}}    \\ 
\bf{Method}                            	& Accuracy      	& Kappa   	& Accuracy   & Kappa	& Accuracy   & Kappa \\ \hline
Full Fine-tuning (\textit{FT})                	&\B85.10 (0.6)	& \B70.20 (1.3)	& 84.59 (1.2)	& 69.19 (2.5)	& 82.83 (1.1)	& 65.65 (2.2)	\\
Linear Probing (\textit{LP})                        & 79.29 (1.6)	& 58.58 (3.3)	& 73.23 (2.1)	& 46.46 (4.3)	& 82.07 (2.8)	& 64.14 (5.6)	\\
Grad. Last $\rightarrow$ First (\textit{G-LF})   	& 84.59 (0.9)	& 69.19 (1.8)	& 84.09 (0.4)	& 68.18 (0.8)	& 83.83 (0.5)	& 67.67 (1.0)	\\
Grad. First $\rightarrow$ Last (\textit{G-FL})    	& 80.81 (1.0)	& 61.62 (2.0)	& 83.58 (1.9)	& 67.17 (3.9)	&\B85.10 (1.7)	& \B70.20 (3.5)	\\
Grad. Last/All (\textit{LP-FT})                   	&\B85.10 (0.2)	& \B70.20 (0.5)	&\B{\U{85.86} (0.6)}	& \B{\U{71.71} (1.3)}	&82.83 (0.2)	& 65.65 (0.5)	\\
$L^1$-$SP$ Regularization                   	& 84.09 (1.1)	& 68.18 (2.3)	& 84.59 (1.2)	& 69.19 (2.5)	& 82.57 (0.7)	& 65.15 (1.5)	\\
$L^2$-$SP$ Regularization                   	& 83.33 (0.4)	& 66.66 (0.8)	& 84.59 (0.2)	& 69.19 (0.5)	& 82.83 (1.1)	& 65.65 (2.2)	\\
Auto-RGN 	                                	& 74.74 (1.9)	& 49.49 (3.9)	& 79.03 (1.2)	& 58.08 (2.5)	& 74.49 (6.3)	& 48.99 (12.7)	\\ \hline
\end{tabular}}
\label{tab:6}
\end{table}

\subsection{MRI}
\begin{figure}[H]
	\centering
	\includegraphics[width=0.9\textwidth]{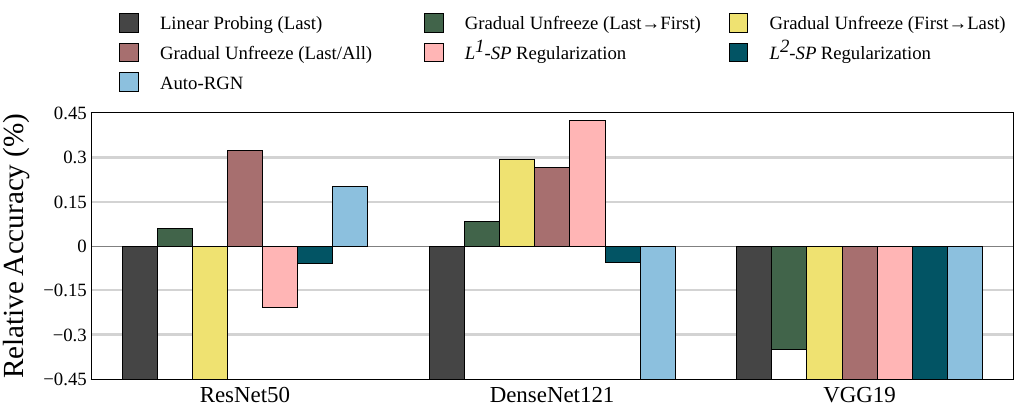}
	\caption{Relative accuracy for T1w MRI, which is calculated as the percentage change in Accuracy compared to Full fine-tuning (\textit{FT}).}
	\label{fig:8}
\end{figure}

The results of our experiments on the T1w MRI dataset, comprising MRI scans of brain tumors, are presented in Table~\ref{tab:7}. We utilize accuracy and F1-score as evaluation metrics. Figure~\ref{fig:8} depicts the relative accuracy improvement over conventional Full Fine-tuning (\textit{FT}). 

\begin{table}[H]
\newrobustcmd{\B}{\bfseries} 
\newrobustcmd{\U}{\underline}
\centering
\caption{Comparison of accuracy and F1-Score along with standard errors across three runs for various fine-tuning methods, evaluated on the T1w MRI dataset. The top scores for each metric and architecture are highlighted in bold, and the overall top score across all architectures is underlined.}
\scalebox{0.9}{
\begin{tabular}{l  c c  c c  c c}
\cline{1-7}
                	& \multicolumn{2}{c}{\bf{ResNet-50}}  	& \multicolumn{2}{c}{\bf{DenseNet-121}} & \multicolumn{2}{c}{\bf{VGG-19}}    \\ 
\bf{Method}                            	& Accuracy      	& F1-Score   	& Accuracy   & F1-Score	& Accuracy   & F1-Score \\ \hline
Full Fine-tuning (\textit{FT})                	& 95.90 (0.6)	& 95.17 (0.7)	& 96.70 (0.5)	& 96.15 (0.5)	&\B{\U{97.29} (0.0)}	& \B96.96 (0.1)	\\
Linear Probing (\textit{LP})                        & 90.32 (0.4)	& 89.20 (0.2)	& 91.58 (0.2)	& 91.48 (0.1)	& 94.37 (0.2)	& 93.29 (0.6)	\\
Grad. Last $\rightarrow$ First (\textit{G-LF})   	& 95.96 (0.6)	& 95.89 (0.8)	& 96.78 (0.2)	& 96.37 (0.1)	& 96.95 (0.3)	& 96.22 (0.3)	\\
Grad. First $\rightarrow$ Last (\textit{G-FL})    	& 92.90 (0.4)	& 91.81 (0.5)	& 96.98 (0.3)	& \B{\U{96.98} (0.1)}	& 96.29 (0.9)	& 95.47 (1.3)	\\
Grad. Last/All (\textit{LP-FT})                  	&\B96.21 (0.2)  & \B95.91 (0.4)	& 96.95 (0.1)	& 96.56 (0.1)	& 96.45 (0.1)	& 95.55 (0.1)	\\
$L^1$-$SP$ Regularization                   	& 95.70 (0.3)	& 95.07 (0.2)	&\B97.11 (0.0)	& 96.52 (0.1)	& 96.41 (0.4)	& 96.09 (0.5)	\\
$L^2$-$SP$ Regularization                   	& 95.85 (0.5)	& 95.25 (0.7)	& 96.64 (0.3)	& 96.12 (0.1)	& 94.87 (0.5)	& 93.71 (0.7)	\\
Auto-RGN 	                                	& 96.10 (0.2)	& 95.48 (0.3)	& 95.59 (1.0)	& 94.97 (1.3)	& 94.46 (0.5)	& 93.12 (0.7)	\\ \hline
\end{tabular}}
\label{tab:7}
\end{table}

Our observations reveal that VGG-19 is the most effective architecture in the MRI domain, followed by DenseNet-121 and ResNet-50. For VGG-19, the baseline \textit{FT} method attains the highest overall accuracy, while all other fine-tuning methods reduce its performance. The \textit{LP-FT} method shows improved accuracy and F1-score with ResNet-50 and DenseNet-121, offering some enhancement over \textit{FT}. Notably, the Gradual Unfreeze methods (\textit{G-LF} and \textit{G-FL}) exhibit varying effects. With ResNet-50 and DenseNet-121, unfreezing layers from last to first (\textit{G-LF}) enhances performance, whereas unfreezing layers from first to last (\textit{G-FL}) has a positive impact only with DenseNet-121 and largely reduces performance with ResNet-50. In the case of regularization methods, only $L^1$-\textit{SP} boosts accuracy for DenseNet-121, while $L^2$-\textit{SP} does not significantly affect outcomes. The Auto-RGN method slightly outperforms \textit{FT} with ResNet-50, but performs worse with the other architectures. Consistently, Linear Probing (\textit{LP}) remains the least effective method, diminishing performance compared to \textit{FT} across all architectures.

\subsection{Histology}

\begin{figure}[H]
	\centering
	\includegraphics[width=0.9\textwidth]{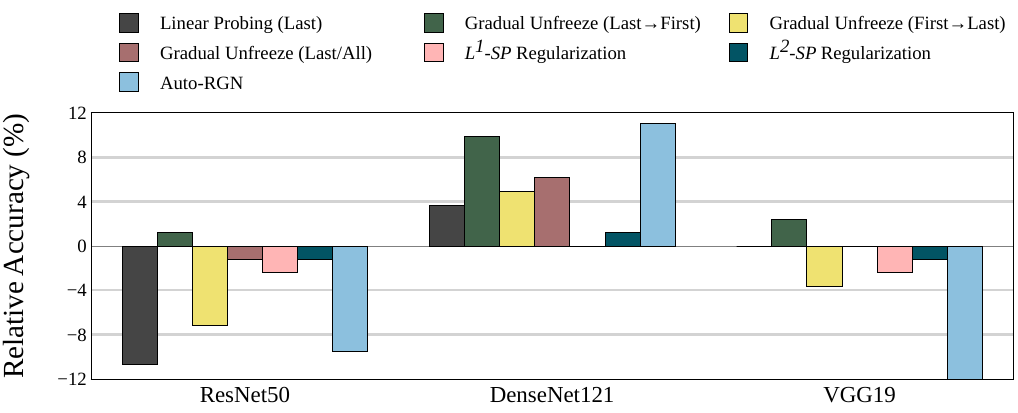}
	\caption{Relative Accuracy for BACH, which is calculated as the percentage change in Accuracy compared to Full fine-tuning (\textit{FT}).}
	\label{fig:9}
\end{figure}

Table~\ref{tab:8} presents the results from our experiments on the BACH dataset, which consists of histopathological images of breast tissue. We utilize two evaluation metrics: accuracy and Kappa. Figure~\ref{fig:9} illustrates the relative accuracy improvement over conventional Full Fine-tuning (\textit{FT}). 

\begin{table}[H]
\newrobustcmd{\B}{\bfseries} 
\newrobustcmd{\U}{\underline}
\centering
\caption{Comparison of accuracy and Kappa along with standard errors across three runs for various fine-tuning methods, evaluated on the BACH dataset. The top scores for each metric and architecture are highlighted in bold, and the overall top score across all architectures is underlined.}
\scalebox{0.9}{
\begin{tabular}{l  c c  c c  c c}
\cline{1-7}
                    & \multicolumn{2}{c}{\bf{ResNet-50}}      & \multicolumn{2}{c}{\bf{DenseNet-121}} & \multicolumn{2}{c}{\bf{VGG-19}} \\
\bf{Method}                               & Accuracy       	& Kappa     	& Accuracy       	& Kappa     	& Accuracy      	& Kappa \\ \hline
Full Fine-tuning (\textit{FT})           	& 70.00 (1.4)	& 60.00 (1.9)	& 67.50 (2.5)	& 56.66 (3.3)	& 69.16 (2.2)	& 58.88 (2.9) \\ 
Linear Probing (\textit{LP})                & 62.50 (3.8)	& 50.00 (5.0)	& 70.00 (2.5)	& 59.99 (3.3)	& 69.16 (0.8)	& 58.89 (1.1) \\ 
Grad. Last $\rightarrow$ First (\textit{G-LF})   &\B70.83 (0.8)	& \B61.11 (1.1)	&74.16 (4.4)	& 65.55 (5.8)	&\B70.83 (4.1)	& \B61.11 (5.5) \\ 
Grad. First $\rightarrow$ Last (\textit{G-FL})   & 65.00 (1.4)	& 53.33 (1.9)	& 70.83 (0.8)	& 61.11 (1.1)	& 66.66 (0.8)	& 55.55 (1.1) \\ 
Grad. Last/All (\textit{LP-FT})               	& 69.16 (1.6)	& 58.89 (2.2)	& 71.66 (3.3)	& 62.22 (4.4)	& 69.16 (0.8)	& 58.89 (1.1) \\ 
$L^1$-$SP$ Regularization               	& 68.33 (1.6)	& 57.77 (2.2)	& 67.50 (1.4)	& 56.66 (1.9)	& 67.50 (3.8)	& 56.66 (5.0) \\ 
$L^2$-$SP$ Regularization               	& 69.16 (0.8)	& 58.89 (1.1)	& 68.33 (2.2)	& 57.77 (2.9)	& 68.33 (1.6)	& 57.77 (2.2) \\ 
Auto-RGN                                	& 63.33 (3.0)	& 51.11 (4.0)	&\B{\U{75.00} (0.0)}	& \B{\U{66.67} (0.0)}	& 52.50 (0.0)	& 36.67 (0.0) \\ \hline
\end{tabular}}
\label{tab:8}
\end{table}
 
Auto-RGN, when paired with DenseNet-121, demonstrated the best performance among all architectures, while the ResNet-50 and VGG-19 achieved identical top accuracy scores. Except for regularization methods, which did not yield significant differences, all the fine-tuning strategies indicated some level of improvement with DenseNet-121. Notably, Auto-RGN exhibited an outstanding accuracy improvement of approximately 10\%, followed by Gradual Unfreeze (\textit{G-LF}) with similar performance. Gradual Unfreeze (\textit{G-FL}) and (\textit{LP-FT}) also yield improvements of around 5\%. For all architectures, only \textit{G-LF} significantly enhanced both accuracy and kappa scores. Surprisingly, Auto-RGN resulted in performance reductions of more than 8\% and 12\% for ResNet-50 and VGG-19, respectively. The limited number of samples (400 images) in this dataset may contribute to such variances in performance.

\subsection{Dermoscopy}

\begin{figure}[H]
	\centering
	\includegraphics[width=0.9\textwidth]{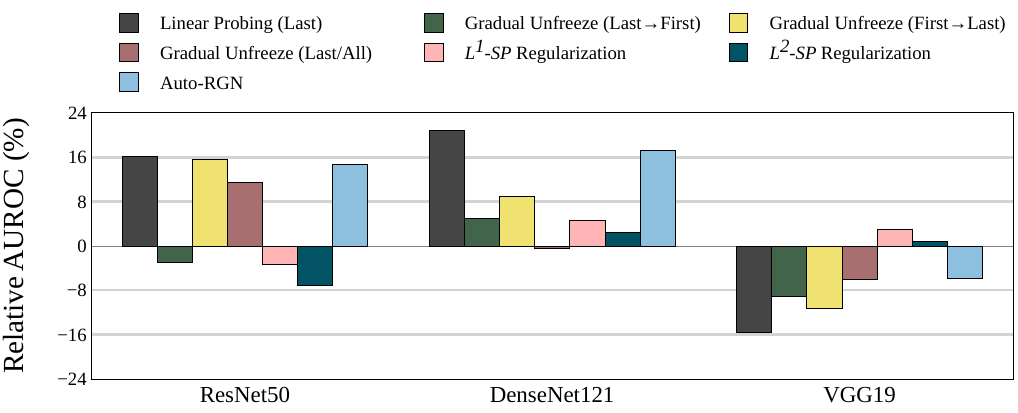}
	\caption{Relative AUROC for ISIC 2020, which is calculated as the percentage change in AUROC compared to Full fine-tuning (\textit{FT}).}
	\label{fig:10}
\end{figure}

Table~\ref{tab:9} presents the results from our experiments on the ISIC 2020 dataset, comprising skin lesion images classified as melanoma or non-melanoma. We employ accuracy and Area Under the Receiver Operating Characteristic Curve (AUROC) as evaluation metrics. Figure~\ref{fig:10} depicts the relative accuracy improvement over conventional Full Fine-tuning (\textit{FT}). 

\begin{table}[H]
\newrobustcmd{\B}{\bfseries} 
\newrobustcmd{\U}{\underline}
\centering
\caption{Comparison of accuracy and AUROC along with standard errors across three runs for various fine-tuning methods, evaluated on the ISIC 2020 dataset. The top scores for each metric and architecture are highlighted in bold, and the overall top score across all architectures is underlined.}
\scalebox{0.9}{
\begin{tabular}{l  c c  c c  c c}
\cline{1-7}
                	   & \multicolumn{2}{c}{\bf{ResNet-50}}  	& \multicolumn{2}{c}{\bf{DenseNet-121}} & \multicolumn{2}{c}{\bf{VGG-19}}    \\ 
\bf{Method}                     & AUROC         & Accuracy              & AUROC & Accuracy & AUROC & Accuracy \\ \hline
Full Fine-tuning (\textit{FT})  & 73.54 (3.6)   &\B{\U{97.98} (0.1)}    & 71.27 (2.9)   & 97.56 (0.1)   & 83.24 (1.5)   & 96.63 (0.6) \\
Linear Probing (\textit{LP})    & \B85.42 (0.2) & 85.61 (0.7)   & \B{\U{86.12} (0.0)}   & 86.47 (0.7)   & 70.23 (0.2)   &\B97.79 (0.0) \\
Grad. Last $\rightarrow$ First (\textit{G-LF})  & 71.28 (1.6)   & 97.40 (0.1)   & 74.76 (2.2)   & 97.52 (0.1)   & 75.61 (1.8) & 97.37 (0.1) \\
Grad. First $\rightarrow$ Last (\textit{G-FL})  & 85.00 (0.3)   & 92.60 (0.5)   & 77.68 (0.9)   & 96.58 (0.2)   & 73.85 (0.1) & 96.82 (0.4) \\
Grad. Last/All (\textit{LP-FT}) & 82.03 (0.6)   & 97.40 (0.3)   & 70.98 (2.3)   & 97.71 (0.0)   & 78.27 (0.5)   & 97.36 (0.2) \\
$L^1$-$SP$ Regularization       & 71.11 (1.8)   & 97.59 (0.2)   & 74.51 (0.8)   & 97.34 (0.2)   & \B85.65 (1.5) & 96.62 (0.9) \\
$L^2$-$SP$ Regularization       & 68.31 (0.4)   & 97.62 (0.0)   & 73.03 (0.4)   &\B{\U{97.98} (0.0)} & 83.90 (2.1)  & 96.73 (0.4) \\
Auto-RGN                        & 84.32 (1.4)   & 97.28 (0.2)   & 83.52 (2.0)   & 97.42 (0.1)   & 78.40 (6.5)   & 76.15 (2.2) \\ \hline
\end{tabular}}
\label{tab:9}
\end{table}

The baseline \textit{FT} achieves the highest accuracy with the ResNet-50 architecture, but registers a lower AUROC. This discrepancy is attributed to the dataset's significant class imbalance, which can render accuracy a less reliable metric. Although accuracy reflects correct classifications, AUROC provides a more comprehensive view of the model’s ability to differentiate between positive and negative cases. Considering the AUROC metric, Linear Probing (\textit{LP}) records the most substantial performance improvements for both ResNet-50 and DenseNet-121, with an increase exceeding 10\%. For VGG-19, \textit{LP} also attains the best accuracy, while the highest AUROC is achieved by $L^1$-\textit{SP}, improving by approximately 3\%. However, for ResNet-50 and DenseNet-121, $L^1$-\textit{SP} reduces performance in both metrics. Auto-RGN does not enhance accuracy in any evaluated model but does provide some AUROC improvement for ResNet-50 and DenseNet-121, showing increases of more than 10\% in both cases.

\subsection{Endoscopic surgery}

\begin{figure}[H]
	\centering
	\includegraphics[width=0.9\textwidth]{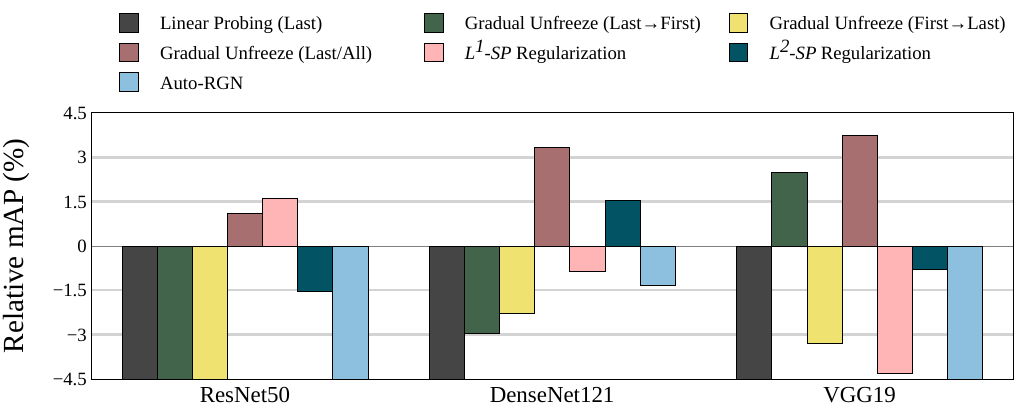}
	\caption{Relative mAP for CholecT50-\textit{Instrument}, which is calculated as the percentage change in Mean Average Precision compared to Full fine-tuning (\textit{FT}).}
	\label{fig:11}
\end{figure}

Table~\ref{tab:10} details the results of our experiments on the CholecT50-\textit{Instrument} dataset, which features endoscopic images of surgical instruments. We utilize mean Average Precision (mAP) and Area Under the Receiver Operating Characteristic Curve (AUROC) as evaluation metrics. Figure~\ref{fig:11} illustrates the relative mAP improvement over conventional Full Fine-tuning (\textit{FT}).

\begin{table}[H]
\newrobustcmd{\B}{\bfseries} 
\newrobustcmd{\U}{\underline}
\centering
\caption{Comparison of mAP and AUROC along with standard errors across three runs for various fine-tuning methods, evaluated on the CholecT50-\textit{Instrument} dataset. The top scores for each metric and architecture are highlighted in bold, and the overall top score across all architectures is underlined.}
\scalebox{0.9}{
\begin{tabular}{l  c c  c c  c c}
\cline{1-7}
                	& \multicolumn{2}{c}{\bf{ResNet-50}}  	& \multicolumn{2}{c}{\bf{DenseNet-121}} & \multicolumn{2}{c}{\bf{VGG-19}}    \\ 
\bf{Method}                            	& mAP      	& AUROC   	& mAP   & AUROC	& mAP   & AUROC \\ \hline
Full Fine-tuning (\textit{FT})                      	& 83.57 (0.1)	& 94.62 (0.1)	& 79.96 (0.2)	& 93.66 (0.1)	& 75.58 (0.4)	& 93.32 (0.0)	\\
Linear Probing (\textit{LP})                         	& 50.21 (0.3)	& 84.59 (0.0)	& 49.74 (0.1)	& 84.49 (0.1)	& 52.47 (0.2)	& 86.39 (0.2)	\\
Grad. Last $\rightarrow$ First (\textit{G-LF})   	& 79.57 (0.4)	& 92.49 (0.6)	& 77.60 (0.4)	& 93.91 (0.4)	& 77.47 (1.4)	& 93.46 (0.1)	\\
Grad. First $\rightarrow$ Last (\textit{G-FL})   	& 74.09 (1.6)	& 91.58 (0.6)	& 78.13 (0.4)	& 92.69 (0.3)	& 73.09 (2.8)	& 92.57 (0.2)	\\
Grad. Last/All (\textit{LP-FT})             &84.50 (0.8)	       & 94.70 (0.2)	&\B82.63 (0.3)	& \B94.21 (0.2)	&\B78.42 (0.6)	& \B93.91 (0.1)	\\
$L^1$-\textit{SP} Regularization            &\B{\U{84.91} (1.1)}	& \B{\U{94.79} (0.4)}	& 79.28 (0.5)	& 93.47 (0.3)	& 72.32 (3.3)	& 92.76 (0.5)	\\
$L^2$-\textit{SP} Regularization            & 82.28 (0.8)	      & 94.13 (0.4)	& 81.21 (0.6)	& 93.48 (0.3)	& 74.98 (0.6)	& 93.02 (0.0)	\\
Auto-RGN                                    & 78.97 (1.1)	      & 93.43 (0.3)	& 78.90 (0.3)	& 93.36 (0.0)	& 56.80 (13.6)	& 83.63 (8.9)	\\\hline
\end{tabular}}
\label{tab:10}
\end{table}

Our observations indicate that Linear Probing (\textit{LP}) underperforms on both metrics across all architectures, suggesting that the last layer of the pre-trained model alone is inadequate for capturing features specific to the endoscopic surgery domain. Among all the architectures and methods evaluated, $L^1$-\textit{SP} with ResNet-50 achieves the highest mAP and AUROC scores, followed by \textit{LP-FT} with DenseNet-121 and VGG-19. \textit{LP-FT} is the only fine-tuning method that consistently improves mAP for all architectures, with VGG-19 and DenseNet-121 exhibiting the most substantial enhancements in mAP over \textit{FT}, exceeding 3\%. The effects of $L^1$-\textit{SP} and $L^2$-\textit{SP} Regularization vary across different architectures. For ResNet-50, $L^1$-\textit{SP} improves mAP by 1\%, while $L^2$-\textit{SP} boosts DenseNet-121 by about 1.5\%. VGG-19 benefits from Gradual Unfreeze Last to First (\textit{G-LF}), which improves performance by over 2\%. However, \textit{G-FL} and Auto-RGN generally perform poorly in this dataset, only reducing performance across all evaluated architectures.

\section{Discussion}
\label{sec:5}
Our study offers insights into the diverse effects of various fine-tuning methods on the efficacy of pre-trained models. A detailed comparison is presented in Table~\ref{tab:11}, showcasing the most effective combinations of fine-tuning strategies and architectures for each specific dataset.

\begin{table}[H]
\newrobustcmd{\B}{\bfseries} 
\newrobustcmd{\U}{\underline}
\centering
\caption{Optimal fine-tuning methods across diverse medical imaging datasets}
\label{tab:11}
\begin{threeparttable}
\begin{tabular}{l c c c c c }
\hline
 & \bf{Medical} & \bf{Model} & \bf{Fine-tuning} & \bf{Performance} \\ 
\multicolumn{1}{c}{\bf{Dataset}} & \bf{domain} & \bf{architecture} & \bf{method} & \bf{improvement (\%)}\textsuperscript{a} \\
\hline
CheXpert                    & X-ray         & ResNet-50     &LP-FT              & 1.9\%\\
MURA-\it{Shoulder}          & X-ray         & VGG-19        &AutoRGN            & 4.7\%\\
MURA-\it{Wrist}             & X-ray         & VGG-19        &$L^1$-\textit{SP}  & 2.2\%\\
MURA-\it{Humerus}           & X-ray         & DenseNet-121  &LP-FT              & 1.5\%\\
T1w MRI                     & MRI           & VGG-19        &FT                 & 0\%\\
BACH            	        & Microscopy    & DenseNet-121  &AutoRGN            & 11.1\%\\
ISIC 2020                   & Dermoscopy    & DenseNet-121  &LP                 & 20.8\%\\
CholecT50-\it{Instrument}   & Endoscopy     & ResNet-50     &$L^1$-\textit{SP}  & 1.6\%\\
\hline
\end{tabular}
\begin{tablenotes}
  \item[a]{Relative performance improvement of main metric over conventional Full fine-tuning (FT) for the corresponding architecture.}
\end{tablenotes}
\end{threeparttable}
\end{table}

The results indicate that no single fine-tuning approach uniformly outperforms others across all datasets. Although Table~\ref{tab:11} serves as a helpful reference for selecting fine-tuning strategies in different contexts, it is important to recognize the considerable variability in performance, indicating that no one combination can be universally optimal. Similar fine-tuning methods may yield close results in one dataset but diverge significantly in another. Therefore, we concentrate our discussion on the robustness of the fine-tuning strategies by examining their performance across all assessed modalities. This provides insights for extension to other medical modalities not included in this study.

Table~\ref{tab:12} summarizes the performance of each fine-tuning method across various architectures and datasets. Linear Probing (\textit{LP}) generally underperforms in most domains, except for the ISIC 2020 dataset, where it shows improvements in AUROC and accuracy. Previous work has highlighted the benefits of \textit{LP} in the presence of a large distribution shift when high-quality pre-trained features are available \cite{kumar22fine}. However, the significant imbalance in the ISIC dataset indicates challenges in discriminating between positive and negative samples, as evidenced by the discrepancy between accuracy and AUROC metrics. Since \textit{LP} only trains the final layer, it could reduce the risk of overfitting to the minority classes, which is a common issue in imbalanced datasets. 

\begin{table}[H]
\newrobustcmd{\B}{\bfseries} 
\newrobustcmd{\U}{\underline}
\centering
\caption{Summary of the number of times the evaluated fine-tuning methods provide performance improvements compared with \textit{FT} ($>0.1$\%) across all experiments and architectures. The effectiveness percentage represents the proportion of experiments with performance improvements relative to the total number of experiments. }
\begin{tabular}{l c c c c c }
\cline{1-6}
\multicolumn{1}{c}{\bf{Method}} & \bf{ResNet-50} & \bf{DenseNet-121} & \bf{VGG-19} & \bf{Total} & \bf{Effectiveness (\%)} \\  \hline
Linear Probing (\textit{LP}) & 0 & 1 & 1 & 2 & 8.3 \\
Grad. Last $\rightarrow$ First (\textit{G-LF}) & 3 & 3 & 5 & 11 & 45.8 \\
Grad. First $\rightarrow$ Last (\textit{G-FL}) & 0 & 3 & 2 & 5 & 20.8 \\
Grad. Last/All (\textit{LP-FT}) & 5 & 7 & 2 & 14 & \B{58.3} \\
$L^1$-\textit{SP} Regularization & 3 & 2 & 2 & 7 & 29.2 \\
$L^2$-\textit{SP} Regularization & 2 & 4 & 1 & 7 & 29.2 \\
Auto-RGN  & 4 & 4 & 2 & 10 & 41.7 \\ \hline
\multicolumn{1}{c}{\bf{Total}} & 17 & \B{24} & 15 &  &  \\ \hline
\end{tabular}
\label{tab:12}
\end{table}

Gradual Unfreezing methods (\textit{G-FL} and \textit{G-LF}) exhibit varied results depending on the architecture and dataset. \textit{G-FL} yields some improvements in smaller datasets like MURA-\textit{Humerus} and BACH, but generally does not significantly enhance performance. By initially unfreezing general layers, the model is prompted to first "forget" general features, potentially impeding its adaptation to task-specific nuances. In contrast, \textit{G-LF}, which begins with LP and transitions to full fine-tuning, shows improvements in 45.8\% of experiments, particularly when combined with VGG-19, underscoring the importance of the direction of unfreezing. This supports previous findings that medical images often suffer from output-level shift and benefit from controlled adaptation, starting with task-specific layers \cite{lee23surgical}. Regularization methods ($L^1$-\textit{SP} and $L^2$-\textit{SP}) yield similar performances across all experiments, showing improvements in about 30\% of them, especially when paired with ResNet-50 and DenseNet-121. Regularization around pre-trained parameters might limit exploration of the fine-tuned models and constrain adaptation to domains significantly divergent from the source domain. Auto-RGN proves effective in 41.7\% of cases, emerging as a useful technique for medical imaging, particularly with ResNet-50 and DenseNet-121, where it achieves top scores in X-ray and Histology domains. Auto-RGN's simplicity, which eliminates the need for multiple training stages and adapts the learning rate dynamically, is a key advantage. Additionally, RGN offers insights into each layer's importance, useful for architecture optimization. Conversely, \textit{LP-FT} performs well across all domains, achieving top scores in nearly all target domains and an overall effectiveness of 58.3\% across all experiments. This method's straightforward approach of alternating between fine-tuning the last layer and all layers seems effective and reliable for most medical imaging domains and can be used as a reference for new setups, although it requires careful determination of the number of steps for alternation.

Among the evaluated architectures, none emerges as universally superior, with ResNet-50, DenseNet-121, and VGG-19 each achieving top scores in three domains. However, our results highlight that different architectures have distinct preferences for fine-tuning methods. ResNet-50 and DenseNet-121, for instance, benefit more from \textit{LP-FT} and Auto-RGN in most domains, accounting for 50\% of all improved cases. Their reduced number of trainable parameters and interconnected block structure seem advantageous for the fine-tuning methods applied. Specifically, DenseNet-121 benefits from fine-tuning strategies in 24 cases, and ResNet-50 in 17. Conversely, VGG-19 favors Gradual Unfreezing (\textit{G-LF}), enhancing performance in five datasets. VGG-19's larger parameter count and lack of skip connections might explain its reduced responsiveness to fine-tuning strategies, as its sequential architecture necessitates features to traverse each block during training, potentially favoring methods with gradual and sequential parameter block unfreezing.

\section{Conclusions}
\label{sec:6}

This study provides a thorough analysis of various fine-tuning methods for pre-trained models across multiple medical image domains. We evaluated eight fine-tuning strategies using three common architectures (ResNet-50, DenseNet-121, VGG-19) across eight datasets representing a range of medical domains (X-ray, MRI, Histology, Dermoscopy, and Endoscopic Surgery). Our findings suggest that the choice of architecture and fine-tuning method must be carefully considered based on the specific attributes and challenges of the target dataset. Notably, different architectures showed preferences for certain fine-tuning strategies, yielding valuable practical insights for method selection in medical image analysis.

Nevertheless, our study has limitations. We limited our focus to a specific set of architectures and methods, potentially overlooking various other combinations. For instance, we did not examine architectures optimized for efficiency or lightweight computation, nor did we investigate fine-tuning methods rooted in knowledge distillation or meta-learning. Moreover, our evaluations relied solely on the original datasets' test sets, which may not comprehensively represent the models' generalization capabilities to unseen data or diverse tasks. Future research could expand this analysis to encompass additional architectures and fine-tuning methods, testing models in more challenging and realistic settings and ablation studies to investigate a wider array of parameter configurations and selection based on specific data characteristics. We aim for our work to offer useful insights and guidelines for choosing appropriate fine-tuning methods for pre-trained models in medical image analysis.

\section*{Acknowledgments}
This work was supported in part by the Japan Science and Technology Agency (JST) CREST including AIP Challenge Program under Grant JPMJCR20D5, and in part by the Japan Society for the Promotion of Science (JSPS) Grants-in-Aid for Scientific Research (KAKENHI) under Grant 22K14221.

\appendix

\bibliographystyle{elsarticle-num} 
\bibliography{biblio}

\begin{thebibliography}{10}
\expandafter\ifx\csname url\endcsname\relax
  \def\url#1{\texttt{#1}}\fi
\expandafter\ifx\csname urlprefix\endcsname\relax\def\urlprefix{URL }\fi
\expandafter\ifx\csname href\endcsname\relax
  \def\href#1#2{#2} \def\path#1{#1}\fi

\bibitem{hu23reinforcement}
M.~Hu, J.~Zhang, L.~Matkovic, T.~Liu, X.~Yang, Reinforcement learning in medical image analysis: Concepts, applications, challenges, and future directions, Journal of Applied Clinical Medical Physics 24~(2) (2023) e13898.
\newblock

\bibitem{shen17deep}
D.~Shen, G.~Wu, H.-I. Suk, Deep learning in medical image analysis, Annual Review of Biomedical Engineering 19~(1) (2017) 221--248, pMID: 28301734.
\newblock

\bibitem{litjens17survey}
G.~Litjens, T.~Kooi, B.~E. Bejnordi, A.~A.~A. Setio, F.~Ciompi, M.~Ghafoorian, J.~A. {van der Laak}, B.~{van Ginneken}, C.~I. Sánchez, A survey on deep learning in medical image analysis, Medical Image Analysis 42 (2017) 60--88.
\newblock

\bibitem{romero20targeted}
M.~Romero, Y.~Interian, T.~Solberg, G.~Valdes, Targeted transfer learning to improve performance in small medical physics datasets, Medical Physics 47~(12) (2020) 6246--6256.
\newblock

\bibitem{pan10survey}
S.~J. Pan, Q.~Yang, A survey on transfer learning, IEEE Transactions on Knowledge and Data Engineering 22~(10) (2010) 1345--1359.
\newblock

\bibitem{hussain19study}
M.~Hussain, J.~J. Bird, D.~R. Faria, A study on cnn transfer learning for image classification, in: A.~Lotfi, H.~Bouchachia, A.~Gegov, C.~Langensiepen, M.~McGinnity (Eds.), Advances in Computational Intelligence Systems, Springer International Publishing, Cham, 2019, pp. 191--202.

\bibitem{kora22transfer}
P.~Kora, C.~P. Ooi, O.~Faust, U.~Raghavendra, A.~Gudigar, W.~Y. Chan, K.~Meenakshi, K.~Swaraja, P.~Plawiak, U.~{Rajendra Acharya}, Transfer learning techniques for medical image analysis: A review, Biocybernetics and Biomedical Engineering 42~(1) (2022) 79--107.
\newblock

\bibitem{kim22transfer}
H.~E. Kim, A.~Cosa-Linan, N.~Santhanam, M.~Jannesari, M.~E. Maros, T.~Ganslandt, Transfer learning for medical image classification: A literature review, BMC medical imaging 22~(1) (2022) 69.

\bibitem{peng22rethinking}
L.~Peng, H.~Liang, G.~Luo, T.~Li, J.~Sun, Rethinking transfer learning for medical image classification (2022).
\newblock \href {http://arxiv.org/abs/2106.05152} {\path{arXiv:2106.05152}}.

\bibitem{sanford20data}
T.~H. Sanford, L.~Zhang, S.~A. Harmon, J.~Sackett, D.~Yang, H.~Roth, Z.~Xu, D.~Kesani, S.~Mehralivand, R.~H. Baroni, T.~Barrett, R.~Girometti, A.~Oto, A.~S. Purysko, S.~Xu, P.~A. Pinto, D.~Xu, B.~J. Wood, P.~L. Choyke, B.~Turkbey, Data augmentation and transfer learning to improve generalizability of an automated prostate segmentation model, American Journal of Roentgenology 215~(6) (2020) 1403--1410.
\newblock

\bibitem{koskinen22automated}
J.~Koskinen, M.~Torkamani-Azar, A.~Hussein, A.~Huotarinen, R.~Bednarik, Automated tool detection with deep learning for monitoring kinematics and eye-hand coordination in microsurgery, Computers in Biology and Medicine 141 (2022) 105121.
\newblock

\bibitem{lavanchy21automation}
J.~L. Lavanchy, J.~Zindel, K.~Kirtac, I.~Twick, E.~Hosgor, D.~Candinas, G.~Beldi, Automation of surgical skill assessment using a three-stage machine learning algorithm, Scientific reports 11~(1) (2021) 5197.

\bibitem{yamada23task}
Y.~Yamada, J.~Colan, A.~Davila, Y.~Hasegawa, Task segmentation based on transition state clustering for surgical robot assistance, in: 2023 8th International Conference on Control and Robotics Engineering (ICCRE), 2023, pp. 260--264.
\newblock

\bibitem{zhang20automatic}
D.~Zhang, Z.~Wu, J.~Chen, A.~Gao, X.~Chen, P.~Li, Z.~Wang, G.~Yang, B.~Lo, G.-Z. Yang, Automatic microsurgical skill assessment based on cross-domain transfer learning, IEEE Robotics and Automation Letters 5~(3) (2020) 4148--4155.
\newblock

\bibitem{manokaran21detection}
J.~Manokaran, F.~Zabihollahy, A.~Hamilton-Wright, E.~Ukwatta, {Detection of COVID-19 from chest x-ray images using transfer learning}, Journal of Medical Imaging 8~(S1) (2021) 017503.
\newblock

\bibitem{wang18deep}
J.~Wang, V.~W. Zheng, Y.~Chen, M.~Huang, Deep transfer learning for cross-domain activity recognition, in: Proceedings of the 3rd International Conference on Crowd Science and Engineering, ICCSE'18, Association for Computing Machinery, New York, NY, USA, 2018.
\newblock

\bibitem{recht19doimagenet}
B.~Recht, R.~Roelofs, L.~Schmidt, V.~Shankar, Do {I}mage{N}et classifiers generalize to {I}mage{N}et?, in: K.~Chaudhuri, R.~Salakhutdinov (Eds.), Proceedings of the 36th International Conference on Machine Learning, Vol.~97 of Proceedings of Machine Learning Research, PMLR, 2019, pp. 5389--5400.

\bibitem{quinonero08dataset}
J.~Quiñonero-Candela, M.~Sugiyama, A.~Schwaighofer, N.~D. Lawrence, Dataset shift in machine learning, Mit Press, Cambridge, MA, USA, 2008.

\bibitem{taori20measuring}
R.~Taori, A.~Dave, V.~Shankar, N.~Carlini, B.~Recht, L.~Schmidt, Measuring robustness to natural distribution shifts in image classification, in: H.~Larochelle, M.~Ranzato, R.~Hadsell, M.~Balcan, H.~Lin (Eds.), Advances in Neural Information Processing Systems, Vol.~33, Curran Associates, Inc., 2020, pp. 18583--18599.

\bibitem{fozilov23endoscope}
K.~Fozilov, J.~Colan, A.~Davila, K.~Misawa, J.~Qiu, Y.~Hayashi, K.~Mori, Y.~Hasegawa, Endoscope automation framework with hierarchical control and interactive perception for multi-tool tracking in minimally invasive surgery, Sensors 23~(24) (2023) 9865.
\newblock

\bibitem{colan23openrst}
J.~Colan, A.~Davila, Y.~Zhu, T.~Aoyama, Y.~Hasegawa, {OpenRST}: An open platform for customizable 3d printed cable-driven robotic surgical tools, IEEE Access 11 (2023) 6092--6105.
\newblock

\bibitem{bhojanapalli21understanding}
S.~Bhojanapalli, A.~Chakrabarti, D.~Glasner, D.~Li, T.~Unterthiner, A.~Veit, Understanding robustness of transformers for image classification, in: Proceedings of the IEEE/CVF International Conference on Computer Vision (ICCV), 2021, pp. 10231--10241.

\bibitem{cai21online}
Z.~Cai, O.~Sener, V.~Koltun, Online continual learning with natural distribution shifts: An empirical study with visual data, in: Proceedings of the IEEE/CVF International Conference on Computer Vision (ICCV), 2021, pp. 8281--8290.

\bibitem{xu22improved}
Z.~Xu, A.~Davila, J.~Wilamowski, S.~Teraguchi, D.~M. Standley, Improved antibody-specific epitope prediction using alphafold and abadapt**, ChemBioChem 23~(18) (2022) e202200303.
\newblock

\bibitem{davila22abadapt}
A.~Davila, Z.~Xu, S.~Li, J.~Rozewicki, J.~Wilamowski, S.~Kotelnikov, D.~Kozakov, S.~Teraguchi, D.~M. Standley, {AbAdapt: an adaptive approach to predicting antibody–antigen complex structures from sequence}, Bioinformatics Advances 2~(1), vbac015 (03 2022).
\newblock

\bibitem{radford21learning}
A.~Radford, J.~W. Kim, C.~Hallacy, A.~Ramesh, G.~Goh, S.~Agarwal, G.~Sastry, A.~Askell, P.~Mishkin, J.~Clark, G.~Krueger, I.~Sutskever, Learning transferable visual models from natural language supervision, in: M.~Meila, T.~Zhang (Eds.), Proceedings of the 38th International Conference on Machine Learning, Vol. 139 of Proceedings of Machine Learning Research, PMLR, 2021, pp. 8748--8763.

\bibitem{peters16causal}
J.~Peters, P.~Bühlmann, N.~Meinshausen, Causal inference by using invariant prediction: identification and confidence intervals, Journal of the Royal Statistical Society. Series B (Statistical Methodology) 78~(5) (2016) 947--1012.

\bibitem{arjovsky19invariant}
M.~Arjovsky, L.~Bottou, I.~Gulrajani, D.~Lopez-Paz, Invariant risk minimization, arXiv preprint arXiv:1907.02893 (2019).

\bibitem{rosenfeld22domain}
E.~Rosenfeld, P.~Ravikumar, A.~Risteski, Domain-adjusted regression or: Erm may already learn features sufficient for out-of-distribution generalization, arXiv preprint arXiv:2202.06856 (2022).

\bibitem{kirichenko22last}
P.~Kirichenko, P.~Izmailov, A.~G. Wilson, Last layer re-training is sufficient for robustness to spurious correlations, arXiv preprint arXiv:2204.02937 (2022).

\bibitem{zhuang21comprehensive}
F.~Zhuang, Z.~Qi, K.~Duan, D.~Xi, Y.~Zhu, H.~Zhu, H.~Xiong, Q.~He, A comprehensive survey on transfer learning, Proceedings of the IEEE 109~(1) (2021) 43--76.
\newblock

\bibitem{shi19deep}
Z.~Shi, H.~Hao, M.~Zhao, Y.~Feng, L.~He, Y.~Wang, K.~Suzuki, A deep cnn based transfer learning method for false positive reduction, Multimedia Tools and Applications 78 (2019) 1017--1033.

\bibitem{nogueira17towards}
K.~Nogueira, O.~A. Penatti, J.~A. {dos Santos}, Towards better exploiting convolutional neural networks for remote sensing scene classification, Pattern Recognition 61 (2017) 539--556.
\newblock

\bibitem{kumar22fine}
A.~Kumar, A.~Raghunathan, R.~Jones, T.~Ma, P.~Liang, Fine-tuning can distort pretrained features and underperform out-of-distribution, arXiv preprint arXiv:2202.10054 (2022).

\bibitem{vrbancic20transfer}
G.~Vrbančič, V.~Podgorelec, Transfer learning with adaptive fine-tuning, IEEE Access 8 (2020) 196197--196211.
\newblock

\bibitem{nagae22automatic}
S.~Nagae, D.~Kanda, S.~Kawai, H.~Nobuhara, Automatic layer selection for transfer learning and quantitative evaluation of layer effectiveness, Neurocomputing 469 (2022) 151--162.
\newblock

\bibitem{li18explicit}
X.~Li, Y.~Grandvalet, F.~Davoine, Explicit inductive bias for transfer learning with convolutional networks, in: J.~Dy, A.~Krause (Eds.), Proceedings of the 35th International Conference on Machine Learning, Vol.~80 of Proceedings of Machine Learning Research, PMLR, 2018, pp. 2825--2834.

\bibitem{howard18universal}
J.~Howard, S.~Ruder, Universal language model fine-tuning for text classification, arXiv preprint arXiv:1801.06146 (2018).

\bibitem{guo19spottune}
Y.~Guo, H.~Shi, A.~Kumar, K.~Grauman, T.~Rosing, R.~Feris, Spottune: Transfer learning through adaptive fine-tuning, in: Proceedings of the IEEE/CVF Conference on Computer Vision and Pattern Recognition (CVPR), 2019.

\bibitem{mukherjee20distilling}
S.~Mukherjee, A.~H. Awadallah, Distilling bert into simple neural networks with unlabeled transfer data (2020).
\newblock \href {http://arxiv.org/abs/1910.01769} {\path{arXiv:1910.01769}}.

\bibitem{davila23gradient}
A.~Davila, J.~Colan, Y.~Hasegawa, Gradient-based fine-tuning strategy for improved transfer learning on surgical images, in: 2023 International Symposium on Micro-NanoMechatronics and Human Science, 2023.

\bibitem{shen21partial}
Z.~Shen, Z.~Liu, J.~Qin, M.~Savvides, K.-T. Cheng, Partial is better than all: Revisiting fine-tuning strategy for few-shot learning, Proceedings of the AAAI Conference on Artificial Intelligence 35~(11) (2021) 9594--9602.
\newblock

\bibitem{ro21autolr}
Y.~Ro, J.~Y. Choi, Autolr: Layer-wise pruning and auto-tuning of learning rates in fine-tuning of deep networks, Proceedings of the AAAI Conference on Artificial Intelligence 35~(3) (2021) 2486--2494.
\newblock

\bibitem{lee23surgical}
Y.~Lee, A.~S. Chen, F.~Tajwar, A.~Kumar, H.~Yao, P.~Liang, C.~Finn, Surgical fine-tuning improves adaptation to distribution shifts (2023).
\newblock \href {http://arxiv.org/abs/2210.11466} {\path{arXiv:2210.11466}}.

\bibitem{kermany18identifying}
D.~S. Kermany, M.~Goldbaum, W.~Cai, C.~C. Valentim, H.~Liang, S.~L. Baxter, A.~McKeown, G.~Yang, X.~Wu, F.~Yan, J.~Dong, M.~K. Prasadha, J.~Pei, M.~Y. Ting, J.~Zhu, C.~Li, S.~Hewett, J.~Dong, I.~Ziyar, A.~Shi, R.~Zhang, L.~Zheng, R.~Hou, W.~Shi, X.~Fu, Y.~Duan, V.~A. Huu, C.~Wen, E.~D. Zhang, C.~L. Zhang, O.~Li, X.~Wang, M.~A. Singer, X.~Sun, J.~Xu, A.~Tafreshi, M.~A. Lewis, H.~Xia, K.~Zhang, Identifying medical diagnoses and treatable diseases by image-based deep learning, Cell 172~(5) (2018) 1122--1131.e9.
\newblock

\bibitem{yadav19deep}
S.~S. Yadav, S.~M. Jadhav, Deep convolutional neural network based medical image classification for disease diagnosis, Journal of Big data 6~(1) (2019) 1--18.

\bibitem{liang20transfer}
G.~Liang, L.~Zheng, A transfer learning method with deep residual network for pediatric pneumonia diagnosis, Computer Methods and Programs in Biomedicine 187 (2020) 104964.
\newblock

\bibitem{chouhan20novel}
V.~Chouhan, S.~K. Singh, A.~Khamparia, D.~Gupta, P.~Tiwari, C.~Moreira, R.~Damaševičius, V.~H.~C. de~Albuquerque, A novel transfer learning based approach for pneumonia detection in chest x-ray images, Applied Sciences 10~(2) (2020).
\newblock

\bibitem{apostolopoulos20covid}
I.~D. Apostolopoulos, T.~A. Mpesiana, Covid-19: automatic detection from x-ray images utilizing transfer learning with convolutional neural networks, Physical and engineering sciences in medicine 43 (2020) 635--640.

\bibitem{maghdid20diagnosing}
H.~S. Maghdid, A.~T. Asaad, K.~Z. Ghafoor, A.~S. Sadiq, M.~K. Khan, Diagnosing covid-19 pneumonia from x-ray and ct images using deep learning and transfer learning algorithms (2020).
\newblock \href {http://arxiv.org/abs/2004.00038} {\path{arXiv:2004.00038}}.

\bibitem{ahuja21deep}
S.~Ahuja, B.~K. Panigrahi, N.~Dey, V.~Rajinikanth, T.~K. Gandhi, Deep transfer learning-based automated detection of covid-19 from lung ct scan slices, Applied Intelligence 51 (2021) 571--585.

\bibitem{basaia19automated}
S.~Basaia, F.~Agosta, L.~Wagner, E.~Canu, G.~Magnani, R.~Santangelo, M.~Filippi, Automated classification of alzheimer's disease and mild cognitive impairment using a single mri and deep neural networks, NeuroImage: Clinical 21 (2019) 101645.
\newblock

\bibitem{oh19classification}
K.~Oh, Y.-C. Chung, K.~W. Kim, W.-S. Kim, I.-S. Oh, Classification and visualization of alzheimer’s disease using volumetric convolutional neural network and transfer learning, Scientific Reports 9~(1) (2019) 18150.

\bibitem{eitel19uncovering}
F.~Eitel, E.~Soehler, J.~Bellmann-Strobl, A.~U. Brandt, K.~Ruprecht, R.~M. Giess, J.~Kuchling, S.~Asseyer, M.~Weygandt, J.-D. Haynes, M.~Scheel, F.~Paul, K.~Ritter, Uncovering convolutional neural network decisions for diagnosing multiple sclerosis on conventional mri using layer-wise relevance propagation, NeuroImage: Clinical 24 (2019) 102003.
\newblock

\bibitem{jonsson19brain}
B.~A. J{\'o}nsson, G.~Bjornsdottir, T.~Thorgeirsson, L.~M. Ellingsen, G.~B. Walters, D.~Gudbjartsson, H.~Stefansson, K.~Stefansson, M.~Ulfarsson, Brain age prediction using deep learning uncovers associated sequence variants, Nature communications 10~(1) (2019) 5409.

\bibitem{naser20brain}
M.~A. Naser, M.~J. Deen, Brain tumor segmentation and grading of lower-grade glioma using deep learning in mri images, Computers in Biology and Medicine 121 (2020) 103758.
\newblock

\bibitem{nawaz18classification}
W.~Nawaz, S.~Ahmed, A.~Tahir, H.~A. Khan, Classification of breast cancer histology images using alexnet, in: Image Analysis and Recognition: 15th International Conference, ICIAR 2018, P{\'o}voa de Varzim, Portugal, June 27--29, 2018, Proceedings 15, Springer, 2018, pp. 869--876.

\bibitem{ferreira18classification}
C.~A. Ferreira, T.~Melo, P.~Sousa, M.~I. Meyer, E.~Shakibapour, P.~Costa, A.~Campilho, Classification of breast cancer histology images through transfer learning using a pre-trained inception resnet v2, in: International conference image analysis and recognition, Springer, 2018, pp. 763--770.

\bibitem{bayramoglu16transfer}
N.~Bayramoglu, J.~Heikkil{\"a}, Transfer learning for cell nuclei classification in histopathology images, in: Computer Vision--ECCV 2016 Workshops: Amsterdam, The Netherlands, October 8-10 and 15-16, 2016, Proceedings, Part III 14, Springer, 2016, pp. 532--539.

\bibitem{vesal18classification}
S.~Vesal, N.~Ravikumar, A.~Davari, S.~Ellmann, A.~Maier, Classification of breast cancer histology images using transfer learning, in: Image Analysis and Recognition: 15th International Conference, ICIAR 2018, P{\'o}voa de Varzim, Portugal, June 27--29, 2018, Proceedings 15, Springer, 2018, pp. 812--819.

\bibitem{mahbod19fusing}
A.~Mahbod, G.~Schaefer, I.~Ellinger, R.~Ecker, A.~Pitiot, C.~Wang, Fusing fine-tuned deep features for skin lesion classification, Computerized Medical Imaging and Graphics 71 (2019) 19--29.
\newblock

\bibitem{mahbod20transfer}
A.~Mahbod, G.~Schaefer, C.~Wang, G.~Dorffner, R.~Ecker, I.~Ellinger, Transfer learning using a multi-scale and multi-network ensemble for skin lesion classification, Computer Methods and Programs in Biomedicine 193 (2020) 105475.
\newblock

\bibitem{mukhlif23incorporating}
A.~A. Mukhlif, B.~Al-Khateeb, M.~A. Mohammed, Incorporating a novel dual transfer learning approach for medical images, Sensors 23~(2) (2023).
\newblock

\bibitem{hasan22dermoexpert}
M.~K. Hasan, M.~T.~E. Elahi, M.~A. Alam, M.~T. Jawad, R.~Martí, Dermoexpert: Skin lesion classification using a hybrid convolutional neural network through segmentation, transfer learning, and augmentation, Informatics in Medicine Unlocked 28 (2022) 100819.
\newblock

\bibitem{spolaor23fine}
N.~Spola{\^o}r, H.~D. Lee, A.~I. Mendes, C.~V. Nogueira, A.~R.~S. Parmezan, W.~S.~R. Takaki, C.~S.~R. Coy, F.~C. Wu, R.~Fonseca-Pinto, Fine-tuning pre-trained neural networks for medical image classification in small clinical datasets, Multimedia Tools and Applications (2023) 1--25.

\bibitem{tajbakhsh16convolutional}
N.~Tajbakhsh, J.~Y. Shin, S.~R. Gurudu, R.~T. Hurst, C.~B. Kendall, M.~B. Gotway, J.~Liang, Convolutional neural networks for medical image analysis: Full training or fine tuning?, IEEE Transactions on Medical Imaging 35~(5) (2016) 1299--1312.
\newblock

\bibitem{kim21new}
Y.~J. Kim, J.~P. Bae, J.-W. Chung, D.~K. Park, K.~G. Kim, Y.~J. Kim, New polyp image classification technique using transfer learning of network-in-network structure in endoscopic images, Scientific Reports 11~(1) (2021) 3605.

\bibitem{patrini20transfer}
I.~Patrini, M.~Ruperti, S.~Moccia, L.~S. Mattos, E.~Frontoni, E.~De~Momi, Transfer learning for informative-frame selection in laryngoscopic videos through learned features, Medical \& Biological Engineering \& Computing 58 (2020) 1225--1238.

\bibitem{liu20finetuning}
X.~Liu, C.~Wang, J.~Bai, G.~Liao, Fine-tuning pre-trained convolutional neural networks for gastric precancerous disease classification on magnification narrow-band imaging images, Neurocomputing 392 (2020) 253--267.
\newblock

\bibitem{jaafari21towards}
J.~Jaafari, S.~Douzi, K.~Douzi, B.~Hssina, Towards more efficient cnn-based surgical tools classification using transfer learning, Journal of Big Data 8 (2021) 1--15.

\bibitem{rajpurkar17chexnet}
P.~Rajpurkar, J.~Irvin, K.~Zhu, B.~Yang, H.~Mehta, T.~Duan, D.~Ding, A.~Bagul, C.~Langlotz, K.~Shpanskaya, M.~P. Lungren, A.~Y. Ng, Chexnet: Radiologist-level pneumonia detection on chest x-rays with deep learning (2017).
\newblock \href {http://arxiv.org/abs/1711.05225} {\path{arXiv:1711.05225}}.

\bibitem{irvin19chexpert}
J.~Irvin, P.~Rajpurkar, M.~Ko, Y.~Yu, S.~Ciurea-Ilcus, C.~Chute, H.~Marklund, B.~Haghgoo, R.~Ball, K.~Shpanskaya, J.~Seekins, D.~A. Mong, S.~S. Halabi, J.~K. Sandberg, R.~Jones, D.~B. Larson, C.~P. Langlotz, B.~N. Patel, M.~P. Lungren, A.~Y. Ng, Chexpert: A large chest radiograph dataset with uncertainty labels and expert comparison, Proceedings of the AAAI Conference on Artificial Intelligence 33~(01) (2019) 590--597.
\newblock

\bibitem{bustos20padchest}
A.~Bustos, A.~Pertusa, J.-M. Salinas, M.~{de la Iglesia-Vayá}, Padchest: A large chest x-ray image dataset with multi-label annotated reports, Medical Image Analysis 66 (2020) 101797.
\newblock

\bibitem{wang23intraclass}
L.~Wang, L.~Zhang, X.~Shu, Z.~Yi, Intra-class consistency and inter-class discrimination feature learning for automatic skin lesion classification, Medical Image Analysis 85 (2023) 102746.
\newblock

\bibitem{cassidy22analysis}
B.~Cassidy, C.~Kendrick, A.~Brodzicki, J.~Jaworek-Korjakowska, M.~H. Yap, Analysis of the isic image datasets: Usage, benchmarks and recommendations, Medical Image Analysis 75 (2022) 102305.
\newblock

\bibitem{yuan21large}
Z.~Yuan, Y.~Yan, M.~Sonka, T.~Yang, Large-scale robust deep auc maximization: A new surrogate loss and empirical studies on medical image classification, in: 2021 IEEE/CVF International Conference on Computer Vision (ICCV), 2021, pp. 3020--3029.
\newblock

\bibitem{rajpurkar17mura}
P.~Rajpurkar, J.~Irvin, A.~Bagul, D.~Ding, T.~Duan, H.~Mehta, B.~Yang, K.~Zhu, D.~Laird, R.~L. Ball, et~al., Mura: Large dataset for abnormality detection in musculoskeletal radiographs, arXiv preprint arXiv:1712.06957 (2017).

\bibitem{chen15enhanced}
J.~Cheng, W.~Huang, S.~Cao, R.~Yang, W.~Yang, Z.~Yun, Z.~Wang, Q.~Feng, Enhanced performance of brain tumor classification via tumor region augmentation and partition, PLOS ONE 10~(10) (2015) 1--13.
\newblock

\bibitem{chen15dataset}
{Cheng, Jun}, {T1-weighted CE-MRI dataset}, [Online]. https://figshare.com/articles/dataset/brain tumor dataset/1512427. Accessed Nov. 13, 2023.

\bibitem{aresta19bach}
G.~Aresta, T.~Araújo, S.~Kwok, S.~S. Chennamsetty, M.~Safwan, V.~Alex, B.~Marami, M.~Prastawa, M.~Chan, M.~Donovan, G.~Fernandez, J.~Zeineh, M.~Kohl, C.~Walz, F.~Ludwig, S.~Braunewell, M.~Baust, Q.~D. Vu, M.~N.~N. To, E.~Kim, J.~T. Kwak, S.~Galal, V.~Sanchez-Freire, N.~Brancati, M.~Frucci, D.~Riccio, Y.~Wang, L.~Sun, K.~Ma, J.~Fang, I.~Kone, L.~Boulmane, A.~Campilho, C.~Eloy, A.~Polónia, P.~Aguiar, Bach: Grand challenge on breast cancer histology images, Medical Image Analysis 56 (2019) 122--139.
\newblock

\bibitem{rotemberg21patient}
V.~Rotemberg, N.~Kurtansky, B.~Betz-Stablein, L.~Caffery, E.~Chousakos, N.~Codella, M.~Combalia, S.~Dusza, P.~Guitera, D.~Gutman, et~al., A patient-centric dataset of images and metadata for identifying melanomas using clinical context, Scientific data 8~(1) (2021) 34.

\bibitem{nwoye23cholectriplet}
C.~I. Nwoye, D.~Alapatt, T.~Yu, A.~Vardazaryan, F.~Xia, Z.~Zhao, T.~Xia, F.~Jia, Y.~Yang, H.~Wang, D.~Yu, G.~Zheng, X.~Duan, N.~Getty, R.~Sanchez-Matilla, M.~Robu, L.~Zhang, H.~Chen, J.~Wang, L.~Wang, B.~Zhang, B.~Gerats, S.~Raviteja, R.~Sathish, R.~Tao, S.~Kondo, W.~Pang, H.~Ren, J.~R. Abbing, M.~H. Sarhan, S.~Bodenstedt, N.~Bhasker, B.~Oliveira, H.~R. Torres, L.~Ling, F.~Gaida, T.~Czempiel, J.~L. Vilaça, P.~Morais, J.~Fonseca, R.~M. Egging, I.~N. Wijma, C.~Qian, G.~Bian, Z.~Li, V.~Balasubramanian, D.~Sheet, I.~Luengo, Y.~Zhu, S.~Ding, J.-A. Aschenbrenner, N.~E. {van der Kar}, M.~Xu, M.~Islam, L.~Seenivasan, A.~Jenke, D.~Stoyanov, D.~Mutter, P.~Mascagni, B.~Seeliger, C.~Gonzalez, N.~Padoy, Cholectriplet2021: A benchmark challenge for surgical action triplet recognition, Medical Image Analysis 86 (2023) 102803.
\newblock

\bibitem{cui18large}
Y.~Cui, Y.~Song, C.~Sun, A.~Howard, S.~Belongie, Large scale fine-grained categorization and domain-specific transfer learning, in: Proceedings of the IEEE Conference on Computer Vision and Pattern Recognition (CVPR), 2018.

\bibitem{cubuk19autoaugment}
E.~D. Cubuk, B.~Zoph, D.~Mané, V.~Vasudevan, Q.~V. Le, Autoaugment: Learning augmentation strategies from data, in: 2019 IEEE/CVF Conference on Computer Vision and Pattern Recognition (CVPR), 2019, pp. 113--123.
\newblock

\bibitem{he16deep}
K.~He, X.~Zhang, S.~Ren, J.~Sun, Deep residual learning for image recognition, in: Proceedings of the IEEE Conference on Computer Vision and Pattern Recognition (CVPR), 2016.

\bibitem{huang17densely}
G.~Huang, Z.~Liu, L.~van~der Maaten, K.~Q. Weinberger, Densely connected convolutional networks, in: Proceedings of the IEEE Conference on Computer Vision and Pattern Recognition (CVPR), 2017.

\bibitem{simonyan15deep}
K.~Simonyan, A.~Zisserman, Very deep convolutional networks for large-scale image recognition (2015).
\newblock \href {http://arxiv.org/abs/1409.1556} {\path{arXiv:1409.1556}}.

\end{thebibliography}





\end{document}